\documentclass[twocolumn, journal]{IEEEtran}  % Comment this line out if you need a4paper

\IEEEoverridecommandlockouts                              % This command is only needed if 
\usepackage{amsbsy}
\usepackage{amstext}
\usepackage{amssymb}
\usepackage{babel}
\usepackage{optidef}
\usepackage{graphics} % for pdf, bitmapped graphics files
\usepackage{epsfig} % for postscript graphics files
\usepackage{times} % assumes new font selection scheme installed
\usepackage{amsmath} % assumes amsmath package installed
\usepackage{cite}
\usepackage{textcomp}
\usepackage{array}
\usepackage{hyperref}
\usepackage{subcaption}
\usepackage{color}
\usepackage[bottom]{footmisc}
\usepackage{subfiles}
\usepackage{multirow}
\usepackage{makecell} 
\usepackage{tabularx,booktabs}
\usepackage{units}
\DeclareMathOperator*{\argmax}{arg\,max}
\DeclareMathOperator*{\argmin}{arg\,min}

\graphicspath{Images/}

\title{Object-Reconstruction-Aware Whole-body Control of Mobile Manipulators 
}

\author{Fatih Dursun, Bruno Vilhena Adorno, Simon Watson, and  Wei Pan % <-this % stops a space
\thanks{*This work was funded by the Republic of Türkiye Ministry
of National Education, Ankara, Türkiye, and by the Royal Academy of Engineering under the Research Chairs and Senior Research Fellowships programme}% <-this % stops a space
\thanks{All authors are with the Manchester Centre for Robotics and AI, University of Manchester, UK. (\emph{for correspondence}: 
        {\tt\small fatih.dursun@postgrad.manchester.ac.uk})}%
}

\begin{document}
\newcommand{\dq}[1]{\underline{\boldsymbol{#1}}}
\newcommand{\quat}[1]{\boldsymbol{#1}}
\newcommand{\mymatrix}[1]{\boldsymbol{#1}} 
\newcommand{\myvec}[1]{\boldsymbol{#1}} 
\newcommand{\mapvec}[2]{\boldsymbol{#1}} 
\newcommand{\crossproduct}[2]{\frac{#1#2-#2#1}{2}} 
\newcommand{\dualvector}[2]{\underline{\boldsymbol{#1}}} 
\newcommand{\dual}{\varepsilon} 
\newcommand{\dotproduct}[2]{\langle#1\rangle} 
\newcommand{\norm}[1]{\left\Vert #1\right\Vert } 
\newcommand{\mydual}[2]{\underline{#1}} 
\newcommand{\hami}[2]{\overset{#1}{\operatorname{\mymatrix H}}} 
\newcommand{\hamidq}[2]{\overset{#1}{\operatorname{\mymatrix H}}_{8}\left(#2\right)} 
\newcommand{\hamilton}[2]{\overset{#1}{\operatorname{\mymatrix H}}\left(#2\right)} 
\newcommand{\hamiquat}[2]{\overset{#1}{\operatorname{\mymatrix H}}_{4}\left(#2\right)} 
\newcommand{\tplus}{\dq{\mathcal{T}}} 
\newcommand{\gett}[1]{\dq{\mathcal{T}}\left(#1\right)} 
\newcommand{\dgett}[1]{\dq{\mathcal{T}}'\left(#1\right)} 
\newcommand{\getp}[1]{\operatorname{\mathcal{P}}\left(#1\right)} 
\newcommand{\dgetp}[1]{\operatorname{\mathcal{P}}'\left(#1\right)} 
\newcommand{\getd}[1]{\operatorname{\mathcal{D}}\left(#1\right)} 
\newcommand{\swap}[1]{\text{swap}\{#1\}} 
\newcommand{\imi}{\hat{\imath}} 
\newcommand{\imj}{\hat{\jmath}} 
\newcommand{\imk}{\hat{k}} 
\newcommand{\real}[1]{\operatorname{\mathrm{Re}}\left(#1\right)} 
\newcommand{\imag}[1]{\operatorname{\mathrm{Im}}\left(#1\right)} 
\newcommand{\imvec}{\boldsymbol{\imath}} 
\newcommand{\dqvector}{\operatorname{vec}} 
\newcommand{\mathpzc}[1]{\fontmathpzc{#1}} 	
\newcommand{\cost}[2]{\underset{\text{#2}}{\operatorname{\text{cost}}}\left(\ensuremath{#1}\right)} 
\newcommand{\diag}[1]{\operatorname{diag}\left(#1\right)} 
\newcommand{\dqframe}[1]{\mathcal{F}_{#1}} 
\newcommand{\ad}[2]{\text{Ad}\left(#1\right)#2} 
\newcommand{\adsharp}[2]{\text{Ad}_{\sharp}\left(#1\right)#2} 
\newcommand{\error}[1]{\tilde{#1}} 
\newcommand{\derror}[1]{\dot{\tilde{#1}}} 
\newcommand{\dderror}[1]{\ddot{\tilde{#1}}} 
\newcommand{\spin}{\text{Spin}(3)} 
\newcommand{\spinr}{\text{Spin}(3){\ltimes}\mathbb{R}^{3}}

\maketitle
%\thispagestyle{empty}
%\pagestyle{empty}
%\bibliographystyle{IEEEtran}

%%%%%%%%%%%%%%%%%%%%%%%%%%%%%%%%%%%%%%%%%%%%%%%%%%%%%%%%%%%%%%%%%%%%%%%%%%%%%%%%

\begin{abstract}

Object reconstruction and inspection tasks play a crucial role in various robotics applications. Identifying paths that reveal the most unknown areas of the object is paramount in this context, as it directly affects reconstruction efficiency. This problem is known as the view path planning problem. Current methods often use sampling-based path planning techniques, evaluating potential views along the path to enhance reconstruction performance. However, these methods are computationally expensive as they require evaluating several candidate views on the path. To this end, we propose a computationally efficient solution that relies on calculating a focus point in the most informative (unknown) region and having the robot maintain this point in the camera field of view along the path. In this way, object reconstruction–related information is incorporated into the whole-body control of a mobile manipulator employing a visibility constraint without the need for an additional path planner. We conducted comprehensive and realistic simulations using a large dataset of 114 diverse objects of varying sizes from 57 categories to compare our method with a sampling-based planning strategy and a strategy that does not employ informative paths using Bayesian data analysis. Furthermore, to demonstrate the applicability and generality of the proposed approach, we conducted real‑world experiments with an 8‑DoF omnidirectional mobile manipulator  and a legged manipulator. Our results suggest that, when compared to a sampling based strategy, there is no statistically significant difference in object reconstruction entropy, and there is a 52.3\% probability that they are practically equivalent in terms of coverage. In contrast, our method is 6.2 to 19.36 times faster in terms of computation time and reduces the total time the robot spends between views by 13.76\% to 27.9\%, depending on the camera FoV and model resolution. When compared with strategies that do not exploit informative paths, our method improves, on average, coverage by 4.9\%  and entropy by 9.72\%  at the expense of spending 8.72\% more time in the reconstruction process.

\end{abstract}

\section{INTRODUCTION}

Autonomous mobile robots have gained significant success in real-world applications such as object reconstruction \cite{Ref:Isler,Ref:Vasquez}, environment exploration \cite{Ref:Bircher}, and inspection \cite{Ref:Naazare}. In these applications, selecting optimal sensor poses, a process known as View Planning Problem (VPP) \cite{Ref:Scott}, is critical because it impacts measurement accuracy, efficiency, and the required time. The optimal viewpoint selection for the object reconstruction task is complex and can be formulated as an NP-Hard Set Covering Problem (SCP) if the object model is available (model-based) \cite{Ref:Landgraf}. The solution to this SCP problem is found offline and generates the minimum number of views required to reconstruct the object. If the model of the object is not available, VPP becomes more challenging and is solved in an iterative and online manner, referred to as Next-Best-View (NBV) estimation.

A common strategy to solve the NBV problem is to use a volumetric representation of the object and ray-casting-based evaluation strategies to score the possible views using a utility function usually called information gain (IG) \cite{Ref:Isler, Ref:Delmerico, Ref:Selin}. In those methods, a set of rays is cast from the camera center within the camera field of view (FoV), and the IG is calculated based on the voxels traversed by the rays.
\begin{figure}[t]
      \centering
      \includegraphics[width=1\linewidth]{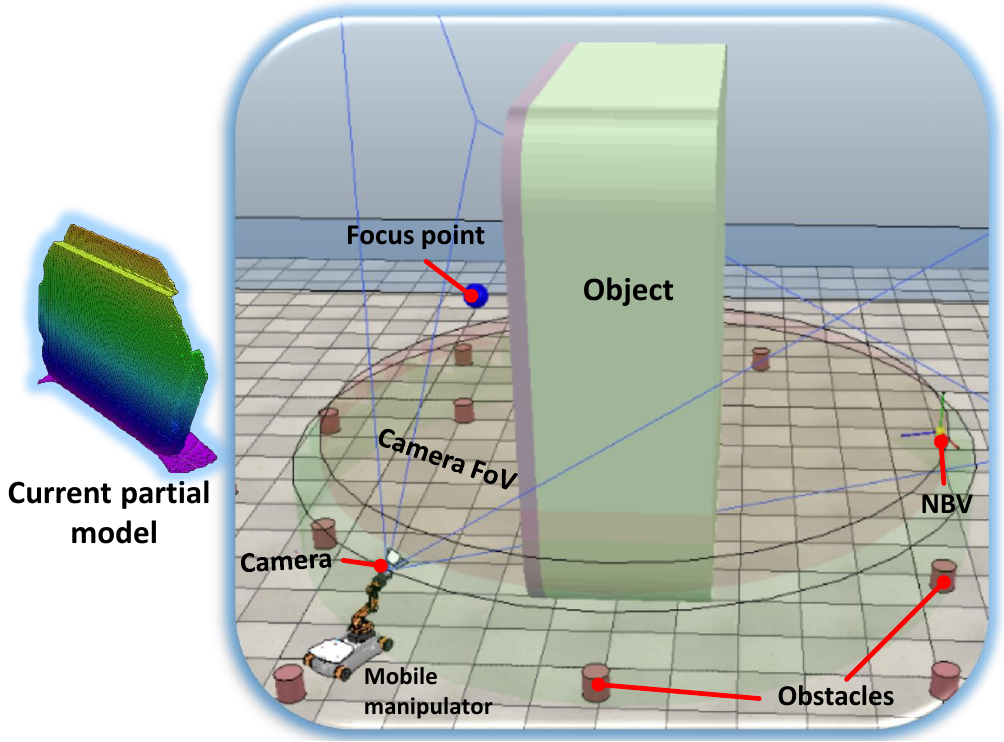}
      \caption{First, an NBV is calculated. Then, the robot moves to the NBV while focusing on a Focus Point and avoiding obstacles in the workspace, enabling the robot to reveal more unknown areas safely.}
      \label{fig:simulation_environment}
\end{figure}
In addition to determining NBVs, the path that the robot follows between them also holds great importance because measurements between NBVs contribute to building the object model and improve the reconstruction process. To solve this problem, referred to as informative path planning (IPP) \cite{Ref:Song_1}, a common approach is to use a sampling-based method to generate candidate views on the path and evaluate them using an IG calculation \cite{Ref:Song_1, Ref:Song_2, Ref:Bircher}. Song \emph{et al.} \cite{Ref:Song_1} proposed a method using micro-aerial vehicles for exploration and inspection tasks. Their algorithm first computes a global path to cover unexplored regions and then plans a local inspection path that scans local frontiers, which are boundary voxels between known and unknown space. Similarly, in \cite{Ref:Song_2}, NBVs that maximize IG are generated, and then a path is planned to the NBV using a sampling-based strategy. In \cite{Ref:Bircher}, an RRT-based candidate view generation strategy has been proposed. Each collision-free node of the tree is evaluated using an IG calculation, which focuses on the unmapped volume that can be explored. Naazare \emph{et al}. \cite{Ref:Naazare} proposed an NBV planner that explores and inspects specific regions of the environment using a mobile manipulator robot. To this end, they proposed a weighted-sum-based utility function that balances the environment exploration and inspection.

Although those methods show significant success, they mainly focus on the exploration problem with a low-resolution representation of the environment. The computation time of these methods becomes exceedingly high when the resolution of the volumetric model is increased, as they require the evaluation of several views on the path using the ray-casting process.
\begin{figure*}[t]
      \centering
      \includegraphics[width=1\linewidth]{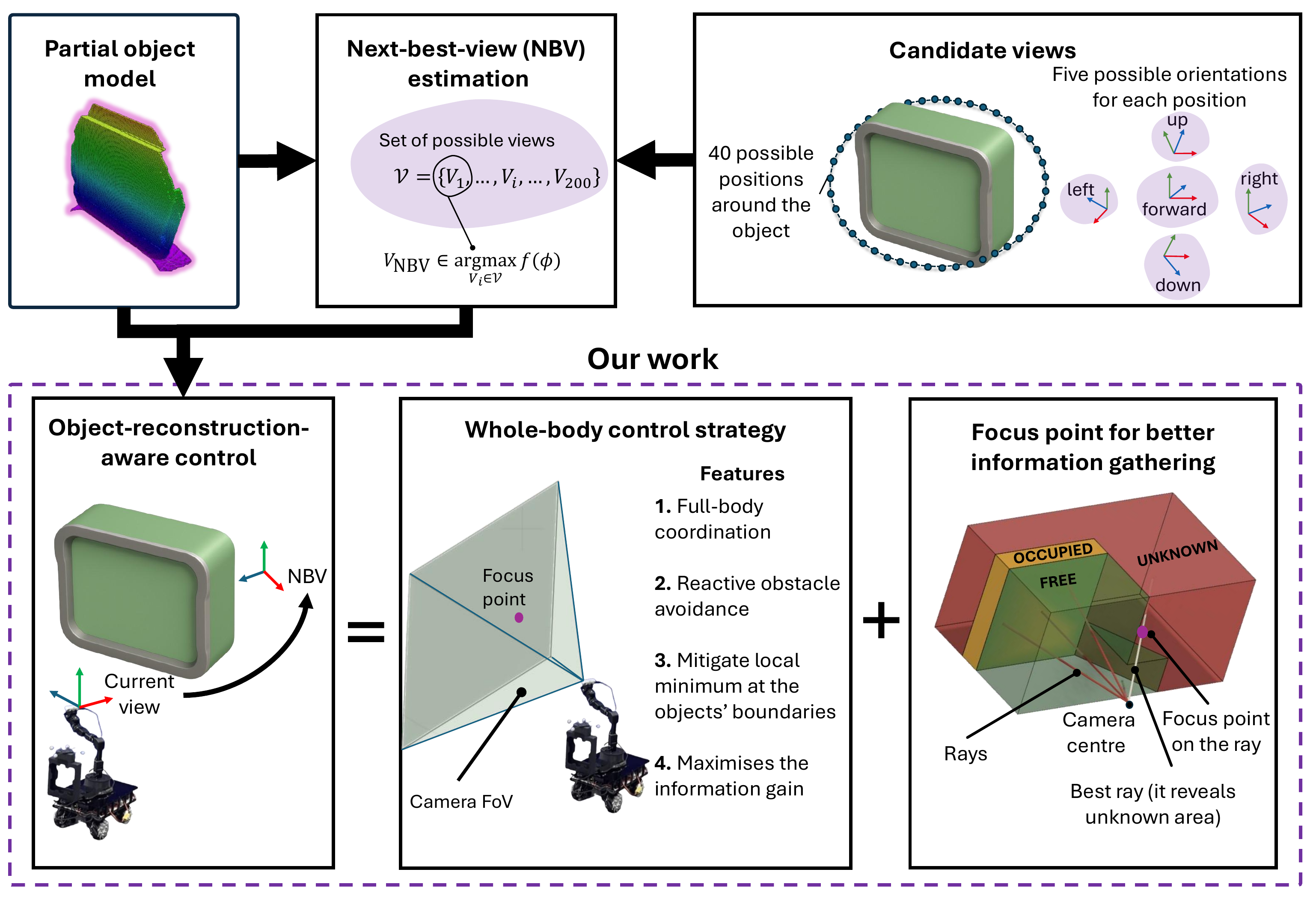}
      \caption{Illustration of the general framework, in which the dashed box highlights our contribution. Given the current partial object model and a set of candidate views, an NBV algorithm (RSV) \cite{Ref:Isler} selects the next view to visit. Our method then uses this target NBV and the current partial model to select an informative focus point and employ a whole-body control strategy to reach the target while keeping the focus point within the camera FoV with a visibility constraint \cite{Ref:Dursun2023}, avoiding obstacles with vector field inequalities (VFIs) \cite{Ref:VFI_Marinho}, and mitigating local minima at obstacle boundaries with a circulation constraint \cite{Ref:Circulation}.}
      \label{fig:general_framework}
\end{figure*}
To mitigate the computational expense of ray-casting-based methods, several works have adopted learning-based strategies for NBV estimation. Mendoza \emph{et al}. \cite{Ref:Mendoza} proposed a supervised learning approach to generate NBVs for 3D reconstruction directly. They generated potential viewpoints around a sphere and used a probabilistic occupancy grid. In their framework, the network takes a 3D volumetric map as input and estimates the best views out of 14 candidate views. Zeng \emph{et al}. \cite{Ref:Zeng} used a point cloud model as the input representation and estimated the information gain for 33 predefined views generated around a sphere encompassing the objects. Han \emph{et al}. \cite{Ref:Han} ranked views directly, rather than first estimating the IG and selecting the view with the highest IG as NBV, since regressing the exact IG for each candidate is considerably more difficult than simply ranking them. Furthermore, Wu \emph{et al}. \cite{Ref:Wu} investigated NBV planning for a specific task, plant phenotyping. In addition to these works, Pan \emph{et al}. \cite{Ref:SCVP} proposed a one-shot learning method where the model outputs the minimum number of views required for complete object coverage among the candidate views instead of estimating a single NBV at each iteration.

Several methods formulate the NBV planning problem within a deep reinforcement learning (DRL) framework. Peralta \emph{et al}. \cite{Ref:Peralta} employed a DRL algorithm to estimate NBVs for house reconstruction using a UAV. In their approach, the current image and the previous image serve as state information (without depth data), and discrete actions are used to drive the robot. Jing \emph{et al}. \cite{Ref:Jing} and Landgraf \emph{et al}. \cite{Ref:Landgraf} applied DRL-based NBV planning for the inspection of industrial objects using a manipulator. Potapova \emph{et al}. \cite{Ref:Potapova} presented a DRL-based approach for the NBV problem applied to arbitrary shapes; here, the discrete action space comprises potential views on a sphere with a fixed stand-off distance. Wang \emph{et al}. \cite{Ref:RLNBV} estimated NBVs using a point cloud model as input and selected NBVs from candidate views as output actions, adopting a strategy similar to Zeng \emph{et al}. \cite{Ref:Zeng} but replacing supervised learning with DRL.

Although learning-based methods tend to be more efficient in terms of computational workload, they are typically trained on objects of specific sizes and rely on a predefined set of candidate views during training. This requires the use of the same candidate views during evaluation, which prevents these methods from being applied to arbitrary views. Consequently, estimating the information gain of an arbitrary view along the path is currently not possible using these approaches.

In this work, instead of focusing on the selection of NBVs, we focus on the path between NBVs to improve object coverage performance. Specifically, we address the question: 
\begin{quote}
\emph{How should the robot move between NBVs to maximize object coverage while overcoming the limitations of ray-casting-based methods, which are computationally expensive, and learning-based methods, which are constrained by specific object sizes and predefined candidate views?}
\end{quote}
Fig.~\ref{fig:general_framework} demonstrates the general framework and highlights the part we focus on in this work.

\begin{figure}[tbh!]
      \centering
      \includegraphics[width=1\linewidth]{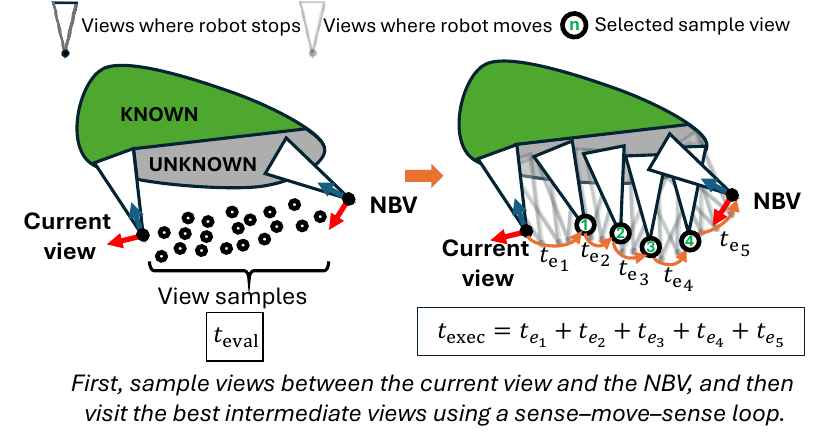}
      \caption{A 2D representation of the general concept adopted by sampling-based IPP strategies.}
      \label{fig:ipp_exec_eval}
\end{figure}

Traditionally, NBV estimation and IPP have been considered as a “sense–move–sense” loop, where sensing is only meaningful at discrete locations and does not consider the continuous robot motion between different views.  Fig.~\ref{fig:ipp_exec_eval} illustrates a 2D example of the general concept of sampling-based methods in the literature that gather information between discrete NBVs: first, many candidate views are sampled along the path to NBV and evaluated to identify the most informative views, then the robot  visits these intermediate views \cite{Ref:Song_1,Ref:Bircher}.

This process involves three main components contributing to planning time: estimating NBVs ($t_\mathrm{nbv}$), sampling and evaluating candidate views ($t_\mathrm{eval}$), and executing the plan by visiting each view ($t_\mathrm{exec}$). Our work aims to reduce the time the robot spends between NBVs ($t_\mathrm{eval}+t_\mathrm{exec}$) by replacing sampling-based view selection and execution with a strategy that leverages continuous, reconstruction-aware robot motion.

To this end, we propose a computationally efficient and generalizable approach. Our method achieves object coverage and entropy performance comparable to state-of-the-art techniques while significantly reducing computation time and eliminating the need for low-resolution model representations. We present a two-stage approach: first, we calculate the NBV from a predefined set of views using the rear-side-voxel method \cite{Ref:Isler}; then, we employ a novel strategy to optimize the path to the NBV (Fig.~\ref{fig:simulation_environment}).

To guide the robot along an informative path, our method calculates a dynamic focus point in the unknown region of the object bounding box. Using a visibility constraint \cite{Ref:Dursun2023}, we instruct the robot to maintain this point in the camera FoV while traveling between NBVs. We used a mobile manipulator for autonomous reconstruction and integrated coverage awareness into the robot control. This integration leverages the mobile manipulator's redundancy and enforces the visibility constraint, eliminating the need for an external path or trajectory planner. Consequently, our method does not evaluate additional candidate views on the local path, making it computationally efficient.

Our approach is simple to implement and can be generalized across various robotic platforms, such as wheeled and legged mobile manipulators.  Furthermore, by leveraging the fact that the camera's roll angle has a minor impact on the volume covered by the camera FoV,  we control only the end-effector position and direction rather than its pose. This task relaxation frees up one degree of freedom, simplifying the task and increasing the robot's functional redundancy.

We conducted comprehensive and realistic simulations with a large dataset of 114 diverse objects of different sizes from the ShapeNet dataset \cite{Ref:Shapenet} to compare our method with a sampling-based planning strategy using Bayesian data analysis. We also conducted real-world experiments using an 8-DOF mobile manipulator and a legged manipulator to demonstrate the successful implementation of the proposed method in a real-world setup.

The main contributions can be summarised as follows:
\begin{itemize}

    \item We introduce a unified framework that integrates the object reconstruction objective into continuous robot control for the IPP problem. Unlike the state-of-the-art sampling methods, the proposed approach does not require sampling and evaluating many candidate views. Instead, it relies on estimating an informative point through an expanded FoV. The method leverages ray-casting to extract information directly from the implicit model while reducing the number of evaluations required. Thus, the proposed method provides a significant reduction in computational time and the total time the robot spends moving between NBVs.

    \item Although NBV estimation relies on a fixed search space, our proposed approach for increasing object coverage while moving between NBVs does not require fixed candidate views along the path for evaluation. In contrast, current learning-based methods do, making our method less restrictive. 
    \item By carefully considering the geometry of the task, we simplify it by freeing up one degree of freedom. To achieve this, we propose a control strategy that considers only the camera position and direction, rather than its pose (i.e., position and orientation). This is done by leveraging the fact that the camera roll angle has a minor impact on the volume covered by the camera's field of view (FoV).

\end{itemize}

\section{Problem Definition}\label{sec:problem_def}
Consider the robot and the object to reconstruct in a 3D Euclidean workspace, $\mathcal{W}\subseteq \mathbb{R}^3$. The object or the area of interest is a closed set of points in the workspace, $\Omega_{\text{obj}}\subset{\mathcal{W}}$, being initially unknown, whereas the location and maximum size are given as prior information. A bounding space $\Omega_{\text{bound}}\subset \mathcal{W}$ contains the object to reconstruct, such that $\Omega_{\text{obj}}\subseteq \Omega_{\text{bound}}$. The interior of the bounding box is categorized as free ($\Omega_{\text{free}}\subseteq\Omega_{\text{bound}}$), occupied ($\Omega_{\text{occ}}\subseteq\Omega_{\text{bound}}$), or unknown ($\Omega_{\text{unknown}}\subseteq\Omega_{\text{bound}}$). In addition, areas that can never be revealed, such as volumes staying behind the occupied region (e.g., the interior of the object) from all possible views, are called residual space, $\Omega_{\text{res}}$. As a result, the bounding box space covers free, occupied, unknown, and residual space, $\Omega_{\text{bound}} = \Omega_{\text{free}} \cup \Omega_{\text{occ}} \cup \Omega_{\text{unknown}} \cup \Omega_{\text{res}}$, where the occupied region represents the surface of object to reconstruct. The goal is to explore all bounding box space except for the residual one, so the problem is considered to be solved when there is no unknown space and, consequently,
\begin{equation}
    \Omega_{\text{bound}} \setminus \Omega_{\text{res}}  = \Omega_{\text{free}}  \cup \Omega_{\text{occ}}.
\end{equation}
%\end{document}

\section{NBV and Informative Focus Point Calculation}{\label{sec:focus_point_calculation}}
\begin{figure}[t]
      \centering
      \includegraphics[width=1\linewidth]{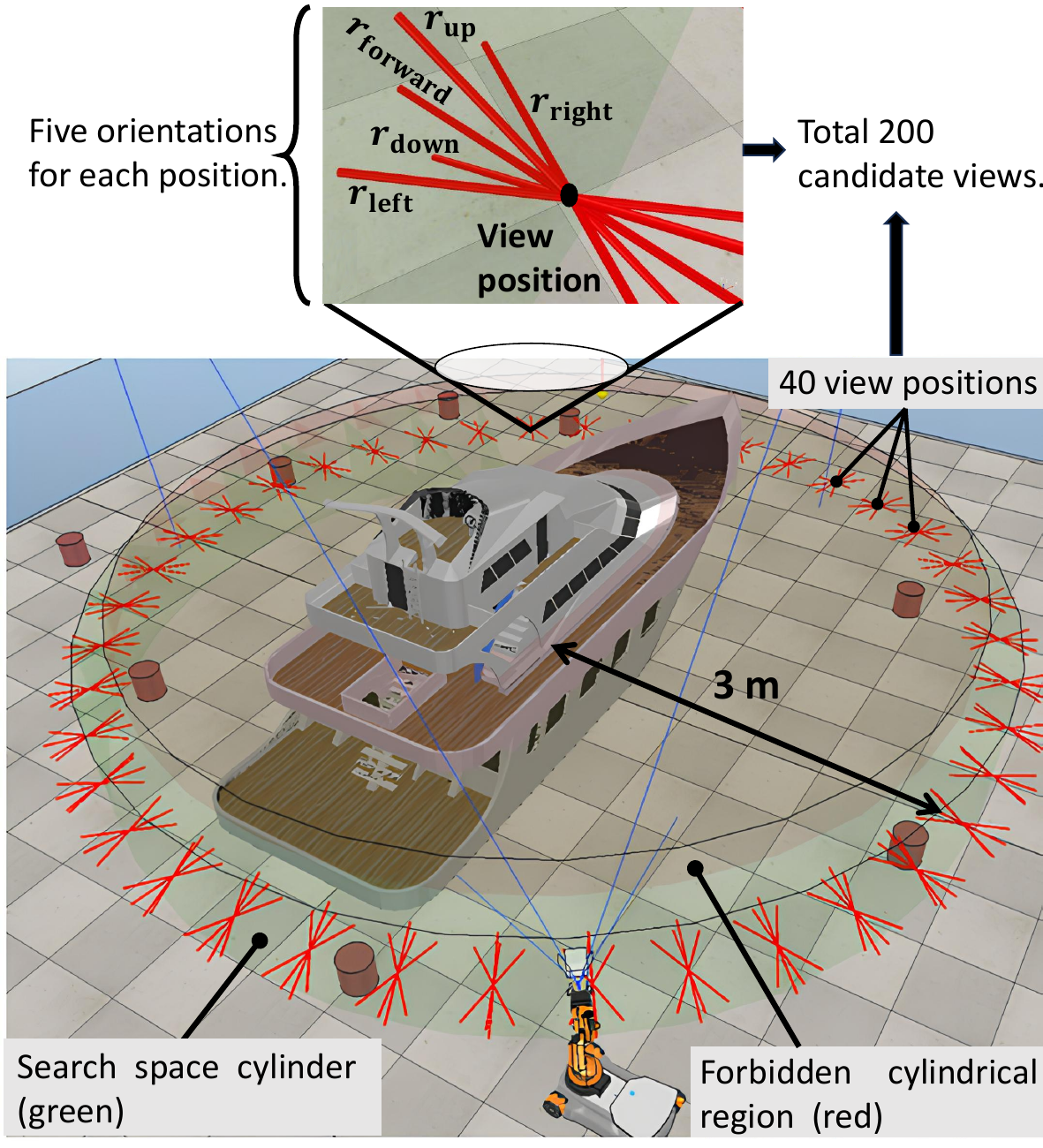}
      \caption{The search space for candidate view generation is defined by 40 positions evenly distributed around a cylinder with a radius of 3~m. Each position has five orientations, resulting in a search space comprising 200 views.}
      \label{fig:search_space}
\end{figure}
\begin{figure*}[t]
      \centering
      \includegraphics[width=1\linewidth]{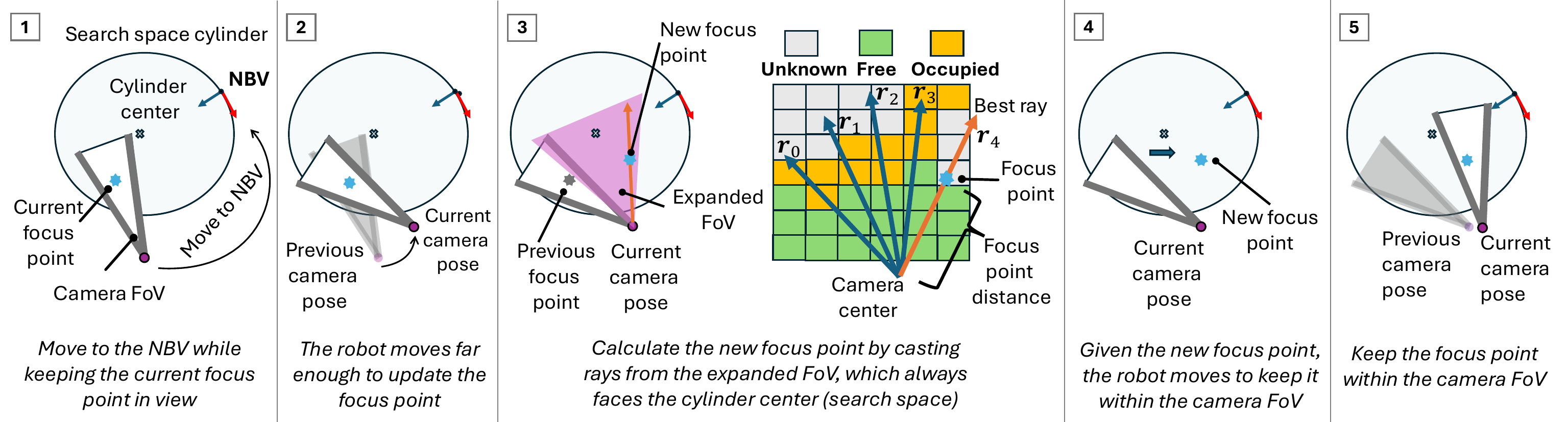}
    
      \caption{Illustration of the stages for focus point calculation. Steps 1–5 are repeated until the robot reaches the NBV.}

      \label{fig:expanded_fov}
\end{figure*}

We propose a strategy that enables the robot to concentrate on informative regions within the object's bounding box. At each iteration, we first determine an NBV using the rear-side-voxel method \cite{Ref:Isler} based on the object's current partial model. It should be noted that although we employed this particular method for NBV estimation, any other NBV algorithm could be used, since the NBV selection is not part of our technical contribution, being used only as a building block in the object 3D reconstruction pipeline. Specifically, our work focuses on improving reconstruction performance \emph{between} NBVs rather than on selecting discrete NBVs. The NBV is selected among the candidate views generated around a vertical cylindrical search space that covers the object, whose center coincides with the center of the bounding box. Selecting views on geometric entities such as spheres, domes, or cylinders is a common approach in the literature \cite{Ref:Isler, Ref:Daudelin, Ref:Han}, as it simplifies the process. Since the robot cannot reach above large and tall objects\textemdash which is the case for the present work\textemdash we used a cylinder as the search space, following a similar methodology from the literature \cite{Ref:Isler}. Without loss of generality, we generate 40 positions evenly distributed around the cylinder, which is dense enough to cover the object but could be adjusted if necessary, and five different orientations at each position (forward, up, down, left, right), forming a search space with 200 candidate views (Fig.~\ref{fig:search_space}). The forward direction is towards the center of the search space, and the angle between the forward and the other four orientations is $\pi/6$ rad. The radius of the search space cylinder is 3~m, and the height from the ground of candidate views is 0.45~m. It is important to emphasise that the cylinder radius and both the height and number of candidate views are parameters of the overall algorithm, which can be easily adjusted to accommodate objects of varying sizes.

Next, we determine the best direction that directs the robot's focus towards the highly informative region (i.e., the region with high entropy) within the object's bounding box. To maintain this direction as the robot moves between NBVs, we calculate the 3D focus point in the selected direction and ensure it stays within the camera FoV using the penultimate constraint in \eqref{eq:control-law}, as detailed in Section~\ref{sec:soft-visibility-constraint}.

To determine an informative direction that helps the robot reveal informative areas, we use ray casting. The steps for calculating a new focus point are illustrated in Fig.~\ref{fig:expanded_fov}. We define a virtual FoV, called the expanded FoV,  that originates at the robot’s current position and always points toward the center of the search-space cylinder, regardless of the camera’s actual orientation. Fig.~\ref{fig:expanded_fov}~(3) illustrates how the virtual FoV faces the cylinder’s center although the real camera’s orientation points to the previous focus point. By doing so, the generated rays are always directed toward the near-object region, since the object is assumed to lie within the search-space cylinder. It should be emphasized that this FoV is purely virtual and exists only to generate candidate rays targeting the near-object area. Furthermore, the FoV’s horizontal and vertical angles determine the volume in which these candidate rays are cast. To increase the chance of hitting informative regions, we select a virtual FoV larger than the camera’s original FoV, thereby expanding the volume in which candidate rays are created. Within the the expanded FoV, we evaluate each ray cast from the camera center using the current object's volumetric model, created using the Octomap library \cite{Ref:Octomap} with a resolution ($r$) of 0.03~m,\footnote{The selection of the resolution depends on the object to be reconstructed. For small-scale objects, a resolution of 0.01 m is commonly used \cite{Ref:Isler}, whereas online inspection of very large objects uses a much lower resolution (larger voxel size), such as 0.3~m \cite{Ref:Song_1}. Since the size of the objects in this work is neither too large nor too small, we opt for a moderate value of 0.03~m.} and select the one with the maximum IG as the best direction. Since the object model is updated with new measurements as the robot moves, we calculate a new direction if the robot's end-effector moves more than a threshold distance from the position where the previous direction was calculated (Fig.~\ref{fig:expanded_fov}~(2)). 

From the current expanded FoV, a set $\mathcal{R}_{v}$ of rays is cast through the current volumetric map $\mathcal{M}_k$, with $k$ being the current reconstruction step. Before hitting the object's surface, a ray $r\in \mathcal{R}_v$ passes through voxels, each adding \emph{volumetric information} \cite{Ref:Isler} to this ray. The ray's total volumetric information, i.e., \emph{ray information gain}, is calculated as
\begin{equation}
    \mathcal{V}_{\mathrm{info}}(r)=\sum_{x\in \mathcal{X}} \mathcal{I}(x), 
\end{equation}
where $\mathcal{X}$ is the set of voxels and $\mathcal{I}(x)$ is volumetric information for voxel $x\in \mathcal{X}$. To focus on unknown regions, we calculate volumetric information based on voxel entropy\cite{Ref:Isler},
\begin{equation}\label{eq:voxel_entropy}
    \mathcal{I}(x)=-P(x)\ln{P(x)}-\overline{P}(x)\ln{\overline{P}}(x),
\end{equation}
where $P(x)$ is the probability of voxel $x$ being occupied and $\overline{P}(x)\triangleq 1 - P(x)$ is the probability of the voxel not being occupied (i.e., it can be free, residual, or unknown). The ray casting stops if it hits an occupied voxel since the volume behind it cannot be revealed. The ray in $\mathcal{R}_{v}$ with the maximum IG (i.e., the optimum) is calculated as
\begin{equation}
\begin{aligned}
r_{\mathrm{opt}}(\mathcal{R}_v)\in \argmax_{r\in \mathcal{R}_v} \quad & \mathcal{V}_\mathrm{info}(r).\\ 
\end{aligned} 
\end{equation}

Fig.~\ref{fig:expanded_fov}~(3) illustrates an example where $r_0$, $r_1$, $r_2$, and $r_3$ traverse free voxels and hit an occupied voxel. There are free and unknown voxels along ray $r_4$. Consequently, this ray would have higher entropy and assist the robot in revealing more unknown areas. On the other hand, rays $r_0$ to $r_3$ hit occupied voxels and therefore have lower entropy. For example, let us consider \eqref{eq:voxel_entropy}, and assume the probabilities of being occupied for unknown and free voxels are 0.5 and 0.1,\footnote{Free voxels could have different probabilities. The voxel is considered free if the probability of being occupied is lower than a certain threshold. A common value is 0.2. See, for instance, \cite{Ref:Isler,Ref:Delmerico}} respectively. In this case, the entropy of $r_2$ would be 0.975, as it encounters three free voxels before reaching an occupied voxel. On the other hand, $r_4$ would have an entropy of 3.054, as it passes through three free and three unknown voxels that are within the maximum ray distance range.\footnote{$\mathcal{I}(x=\mathrm{free})=-0.1\ln(0.1)-0.9\ln(0.9)=0.325$ and $\mathcal{I}(x=\mathrm{unknown})=-0.5\ln(0.5)-0.5\ln(0.5)=0.693$.}

After determining the best direction given by $r_{\mathrm{opt}}(\mathcal{R}_v)$, an updated focus point is selected at a fixed distance from the camera center along this direction. The main idea is that by keeping this point within the camera FoV, the robot reveals informative regions, since the point lies along an informative ray. Consequently, the robot’s overall goal is to move toward the selected NBV while maintaining the chosen focus point in its camera FoV. It is important to distinguish between the expanded FoV, which is a virtual search space used only to generate candidate rays toward the near-object area, and the real FoV of the camera mounted on the robot’s end effector. Although the focus point is computed using the expanded FoV, the robot should keep this point inside its real camera FoV to reveal the informative area (including the focus point), because the measurement region is defined by the real FoV.

\begin{figure}[t]
      \centering
      \includegraphics[width=1\linewidth]{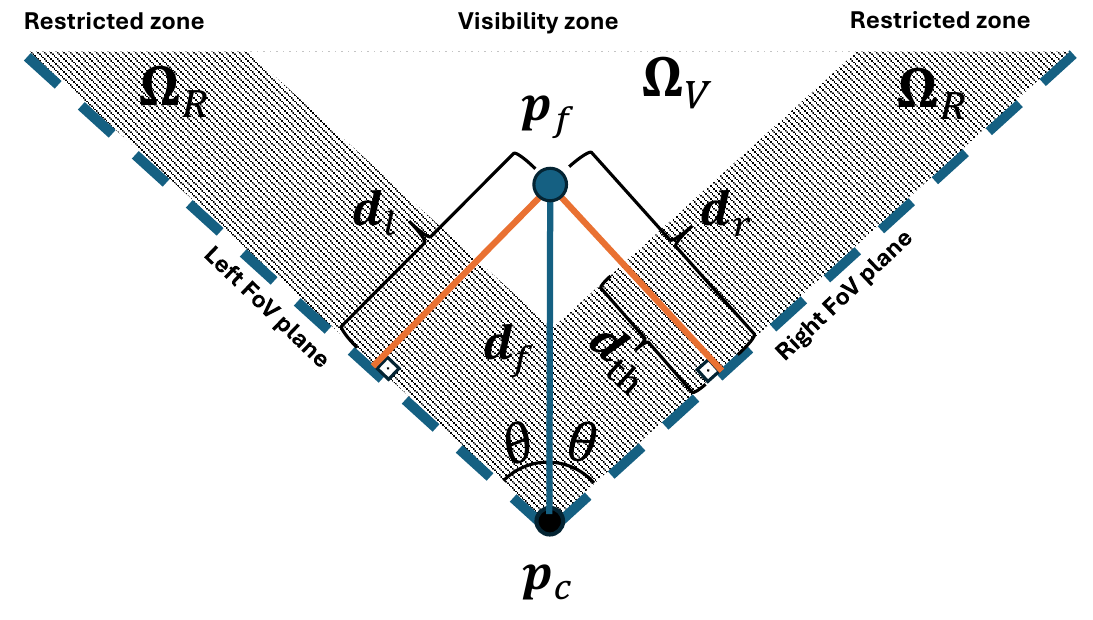}
      \caption{Top view of the geometrical relationship between the focus point and the angle of the camera FoV. The restricted zone ($\Omega_R$) and the visibility zone ($\Omega_V$) are represented with gray and white, respectively, whereas blue dashed lines are the  left and right FoV planes. Furthermore, $\myvec{p}_f$ is the focus point and $\myvec{p}_c$ is the origin of the camera FoV.}
      \label{fig:focus_point_selection}
\end{figure}

To ensure the focus point remains within the camera FoV, we use the visibility‑constraint approach proposed in \cite{Ref:Dursun2023}, as detailed in Section~\ref{sec:soft-visibility-constraint}. The visibility constraint keeps the focus point inside the visibility zone ($\Omega_V$) by ensuring that the distances ($d_l,d_r$) between the FoV planes and the focus point are higher than a threshold ($d_{\mathrm{th}}$), as shown in Fig.~\ref{fig:focus_point_selection}. The figure illustrates from the top view how the visibility and restricted zones are formed by the left and right FoV planes (represented by the blue dashed lines) and the threshold distance ($d_{\mathrm{th}}$). Notice that although the focus point ($p_f$)  is placed on the center line for clear illustration of the related distances, it may lie anywhere within the visibility zone without violating the visibility constraint. Specifically, the constraint does not require the robot to align the camera’s center line with $p_f$; it only ensures that $p_f$ remains within the visibility zone. This relaxation provides the robot with more freedom to drive the end-effector to the target position and direction while still focusing on the visibility region.

The size of the visibility region (thus the degree of relaxation) is determined by the threshold distance ($d_{\mathrm{th}}$): the greater the threshold distance, the more restrictive the visibility constraint. Furthermore, the focus point can be arbitrary on the ray, but the threshold distance must be adjusted accordingly. For example, if the focus point’s distance to the camera center is chosen to be small while the threshold distance is comparatively large, the resulting visibility constraint will be very tight. In our implementation,  $d_f=2.5~\mathrm{m}$, and $d_\mathrm{th}=0.75~\mathrm{m}$, which is 51\% of $d_r$ because the camera FoV angle ($2\theta$) is $74^{\circ}$ (Table~\ref{tab:parameters}).

\subsection{Computational Complexity Analysis}
The main difference between sampling-based IPP methods and the focus point method is that sampling-based methods sample several candidate views along the path and evaluate the sampled views to find the best intermediate view to visit. In contrast, our method calculates focus points in an expanded FoV.

To perform a comparative computational complexity analysis between the two methods, let us assume that the 
sampling-based methods visits $N_\mathrm{sampling}$ intermediate views, while our method calculates $N_\mathrm{focus}$ different focus points as the robot transitions from its current viewpoint to the NBV. This analysis does not include the robot motion, as the goal is to demonstrate the computational complexity of calculating views along the path between NBVs for sampling-based methods, and calculating focus points for our method.

At each step, sampling-based methods evaluate $K$ candidate views and select the best one, whereas our method computes a single focus point from a single expanded FoV. Both approaches rely on ray casting with identical angular resolution $\varsigma$, octomap resolution $r_\mathrm{res}$, and maximum ray length $d_{\max}$.

\subsubsection{Sampling-based methods}
Let the camera FoV be defined as $\mathrm{F}_s \triangleq (\mathrm{F}_h, \mathrm{F}_v)$ with $\mathrm{F}_h$ and $\mathrm{F}_v$ being the horizontal and vertical FoV angles, respectively. The number of rays cast per view is given by
\begin{equation}
R_h =\frac{\mathrm{F}_h}{\varsigma}+1, \quad
R_v = \frac{\mathrm{F}_v}{\varsigma}+1, \quad
R_s \triangleq R_h R_v,
\end{equation}
where $R_h,~R_v$ and $R_s$ are the horizontal, vertical and total number of rays, respectively.
A ray with maximum length of $d_{\max}$ intersects at approximately $V_{r_\mathrm{res}} \approx (d_{\max}/r_\mathrm{res})\max(\cos(\alpha),\sin(\alpha))$ voxels, where $\alpha$ is the ray angle. Querying voxel occupancy along a ray in an octree has logarithmic complexity with respect to the number of voxels \cite{Ref:Bircher}, resulting in a per-ray complexity of $\mathcal{O}(\log V_{r_\mathrm{res}})$. Thus, the computational complexity of evaluating a single candidate view is $\mathcal{O}(R_s \log V_{r_\mathrm{res}})$. At each intermediate step, $K$ candidate views are evaluated, yielding a per-step complexity of $\mathcal{O}(K R_s \log V_{r_\mathrm{res})}$. Considering $N_\mathrm{sampling}$ intermediate steps, the total computational complexity of sampling-based IPP methods is
\begin{equation}
    \mathcal{O}(N_\mathrm{sampling} K R_s \log V_{r_\mathrm{res}}).
\end{equation}

\subsubsection{Focus point method}\label{subsec:focus_point_method}
The proposed method evaluates focus points using an expanded FoV defined as $F_f \triangleq (a_h\mathrm{F}_h,\,a_v\mathrm{F}_v)$, where $a_h,a_v\in (1,\infty)$. The corresponding number of rays evaluated per focus point is $R_f = (a_hF_h/\varsigma+1)\cdot(a_vF_v/\varsigma+1)$. Using the same octree query model, the computational cost of estimating a single focus point is $\mathcal{O}(R_f\log V_{r_\mathrm{res}})$. Since the focus point is computed $N_\mathrm{focus}$ times,\footnote{Recall that a new focus point is calculated along the informative path each time the robot’s end-effector moves more than a threshold distance from the position where the previous direction was calculated. Therefore, $N_\mathrm{focus}$ is approximately the distance between consecutive NBVs divided by that threshold.} the total computational complexity of the proposed focus point method is
\begin{equation}
\mathcal{O}(N_\mathrm{focus}R_f \log V_{r_\mathrm{res}})=\mathcal{O}( a_h a_v N_\mathrm{focus}R_s\log V_{r_\mathrm{res}}),
\end{equation}
as $R_f \approx a_ha_vR_s.$
Sampling-based IPP methods evaluate $K$ sample views to select the best view for $N_\mathrm{sampling}$ times along the path to the NBV. In contrast, the proposed focus point estimation method evaluates a single view with expanded camera FoV to select a focus point for $N_\mathrm{focus}$ times along the path to the NBV. Thus, whenever
\begin{equation}
\label{eq:complexity_relation}
KN_\mathrm{sampling} > a_h a_vN_\mathrm{focus},
\end{equation}
sampling-based IPP methods are computationally more expensive than our proposed approach. Note that $KN_\mathrm{sampling}$ represents the total number of evaluated view samples until reaching the NBV using sampling-based methods. In our setting, we choose $F_s = (74^\circ,60^\circ)$, and $F_f = (90^\circ, 90^\circ)$. Thus, $a_h\approx1.22$ and $a_v = 1.5$, so as long as the number of all evaluated view samples ($KN_\mathrm{sampling}$) is greater than $1.83N_\mathrm{focus}$, our method is computationally more efficient.

\section{Visibility-Aware Kinematic Control}{\label{sec2}}
The relation between configuration-space velocities and task-space velocities of a serial mobile manipulator is given by the differential kinematics equation,
\begin{equation}
    \dot{\myvec{x}}=\mymatrix{J}\dot{\myvec{q}},  
\end{equation}
where  $\mymatrix{J}\triangleq \mymatrix{J}(\myvec{q}) \in \mathbb{R}^{m\times n}$ is the task Jacobian matrix, with $\myvec{q}=[\myvec{q}_\mathrm{base}^T \quad \myvec{q}_\mathrm{arm}^T]^T \in\mathbb{R}^{n}$ being the configuration vector containing the base configuration $\myvec{q}_\mathrm{base} \in\mathbb{R}^{n_b} $ and arm configuration $\myvec{q}_\mathrm{arm} \in\mathbb{R}^{n_a}$, and $\myvec{x}\triangleq \myvec{x}(\myvec{q}) \in \mathbb{R}^m$ is the task-space vector.

Our primary objective is to drive the robot end-effector to a desired position and direction. To this end, we define the task-space variable $\myvec{x}=[\myvec{p}^T \quad \myvec{l}^T]^T \in\mathbb{R}^{6}$ comprising the robot end-effector position ($\myvec{p}\in\mathbb{R}^3$) and direction ($\myvec{l}\in\mathbb{R}^3$) vectors, with corresponding task  Jacobian given by $\mymatrix{J}=[\mymatrix{J}_p \quad \mymatrix{J}_l]^T \in\mathbb{R}^{6\times n}$ with $\mymatrix{J}_p \triangleq \mymatrix{J}_p(\myvec{q}) \in \mathbb{R}^{3\times n}$ and $\mymatrix{J}_l \triangleq \mymatrix{J}_l(\myvec{q}) \in \mathbb{R}^{3\times n}$ being translation and line-direction Jacobians \cite{Ref:VFI_Marinho}, respectively.

Given the constant target task-space vector $\myvec{x}_d \in\mathbb{R}^{6}$, with  $\dot{\myvec{x}}_{d}=\myvec{0}$ for all $t$, and corresponding error $\error{\myvec{x}}=\myvec{x}-\myvec{x}_d$, a motion control law $\myvec{u} \in \mathbb{R}^n$ that enforces an exponential error decay as best as possible while ensuring collision avoidance \cite{Ref:Marinho} is given by
\begin{equation}
\label{eq:control-law}
\begin{aligned}
\left(\myvec{u},\myvec{\kappa}\right) \in \argmin_{\dot{\myvec{q}},\myvec{\kappa}} \  & \norm{\myvec{J}\dot{\myvec{q}}+\lambda \myvec{\error{x}}}_{2}^2+
\lambda_q\norm{\dot{\myvec{q}}}_{2}^{2}+\lambda_{\kappa}\norm{\myvec{\kappa}}_{2}^2 \\
\textrm{subject to} \  &-\nabla \error{D}\left(\myvec{q}\right)^T\dot{\myvec{q}}\leq \lambda_e\error{D}(\myvec{q})
\\&  -\nabla \error{D}_b\left(\myvec{q}_{\mathrm{base}}\right)^T\dot{\myvec{q}}_{\mathrm{base}}\leq \lambda_d\error{D}_b(\myvec{q}_{\mathrm{base}})
\\& -\myvec{T}(\nabla \error{D}_b)^T\dot{\myvec{q}}_\mathrm{base}\leq -\beta(\error{D}_b(\myvec{q}_{\mathrm{base}}))
\\&-\dot{\myvec{q}}_{\mathrm{lim}}\preceq\dot{\myvec{q}}\preceq \dot{\myvec{q}}_{\mathrm{lim}}
\\&-\lambda_\phi\error{\myvec{q}}_{\mathrm{arm}_l} \preceq\dot{\myvec{q}}_\mathrm{arm}\preceq -\lambda_\phi\error{\myvec{q}}_{\mathrm{arm}_u}
\\& -\mymatrix{J}_v(\myvec{q}) \dot{\myvec{q}}-\myvec{\kappa} \preceq \lambda_v\error{\myvec{d}}_v(\myvec{q})
\\& \,\,\myvec{\kappa}\succeq \myvec{0},
\end{aligned} 
\end{equation}
where $\nabla \error{D}(\myvec{q}) \in \mathbb{R}^n$ and $\nabla \error{D}_b(\myvec{q}_\mathrm{base}) \in \mathbb{R}^{n_b}$ are the gradient of $\error{D}(\myvec{q})\in\mathbb{R}$ and $\error{D}_b(\myvec{q}_\mathrm{base})\in\mathbb{R}$, which are 
smooth soft minimum distance functions for the end-effector and robot base respectively, further described in \eqref{eq:softmin}. Constants $\lambda_d\in\left(0,\infty \right)$ and $\lambda_b\in\left(0,\infty \right)$ determine how quickly the robot is allowed to approach obstacles in the workspace. Furthermore, 
$ \mymatrix{T}(\nabla \error{D}_b(\myvec{q}_\mathrm{base})) \in \mathbb{R}^{n_b}$ is the tangent vector to $\nabla \error{D}_b(\myvec{q}_\mathrm{base})$, and $\beta(\error{D}_b(\myvec{q}_{\mathrm{base}}))$ is a distance-dependent continuous function that activates circulation when the robot is close to the obstacle (see ~\ref{subsubsec:circulation} for more details on $\mymatrix{T}$ and $\beta$). Additionally, $\myvec{0} \preceq \dot{\myvec{q}}_{\mathrm{lim}} \in \mathbb{R}^n$ is the bound for the configuration velocities, whereas $\error{\myvec{q}}_{\mathrm{arm}_u}\triangleq\myvec{q}_\mathrm{arm}-\myvec{q}_u$ and $\error{\myvec{q}}_{\mathrm{arm}_l}\triangleq\myvec{q}_\mathrm{arm}-\myvec{q}_l$, with $\myvec{q}_u,\myvec{q}_l \in (-\pi,\pi)^{n_a}$ are used to enforce the upper and lower joint limits for the arm, and $\lambda_\phi \in\left(0,\infty \right)$ being defined to enforce the allowable approach velocity to the joint limits. Finally, $\mymatrix{J}_v(\myvec{q})=\partial\error{\myvec{d}}_v/\partial\myvec{q}  \in\mathbb{R}^{{n_v}\times n}$, $\error{\myvec{d}}_v\triangleq \error{\myvec{d}}_v(\myvec{q}) \in\mathbb{R}^{n_v}$, and $\lambda_v\in\left(0,\infty \right)$ are the Jacobian matrix, distance vector, and gain for the visibility constraint, respectively, and $\myvec{\kappa}\in\mathbb{R}^{n_v}$ is the slack variable vector with corresponding weight given by $\lambda_{\kappa}\in\left(0,\infty \right)$. The first two constraints will be further detailed in Subsection~\ref{subsec:collision_avoidance}.

The first term in the objective function of control law \eqref{eq:control-law} determines the desired closed-loop task-space error dynamics (i.e., exponential stabilization), the second term prevents large joint velocities, and the third term minimizes the slack variable vector, $\myvec{\kappa}$, ensuring the soft visibility constraint will be respected as best as possible, but without overly affecting the primary objective of driving the end-effector to the desired target.
\subsection{Task-Jacobian and Desired Closed-Loop Task-Error Dynamics}
\begin{figure}[h]
      \centering
      \includegraphics[width=1\linewidth]{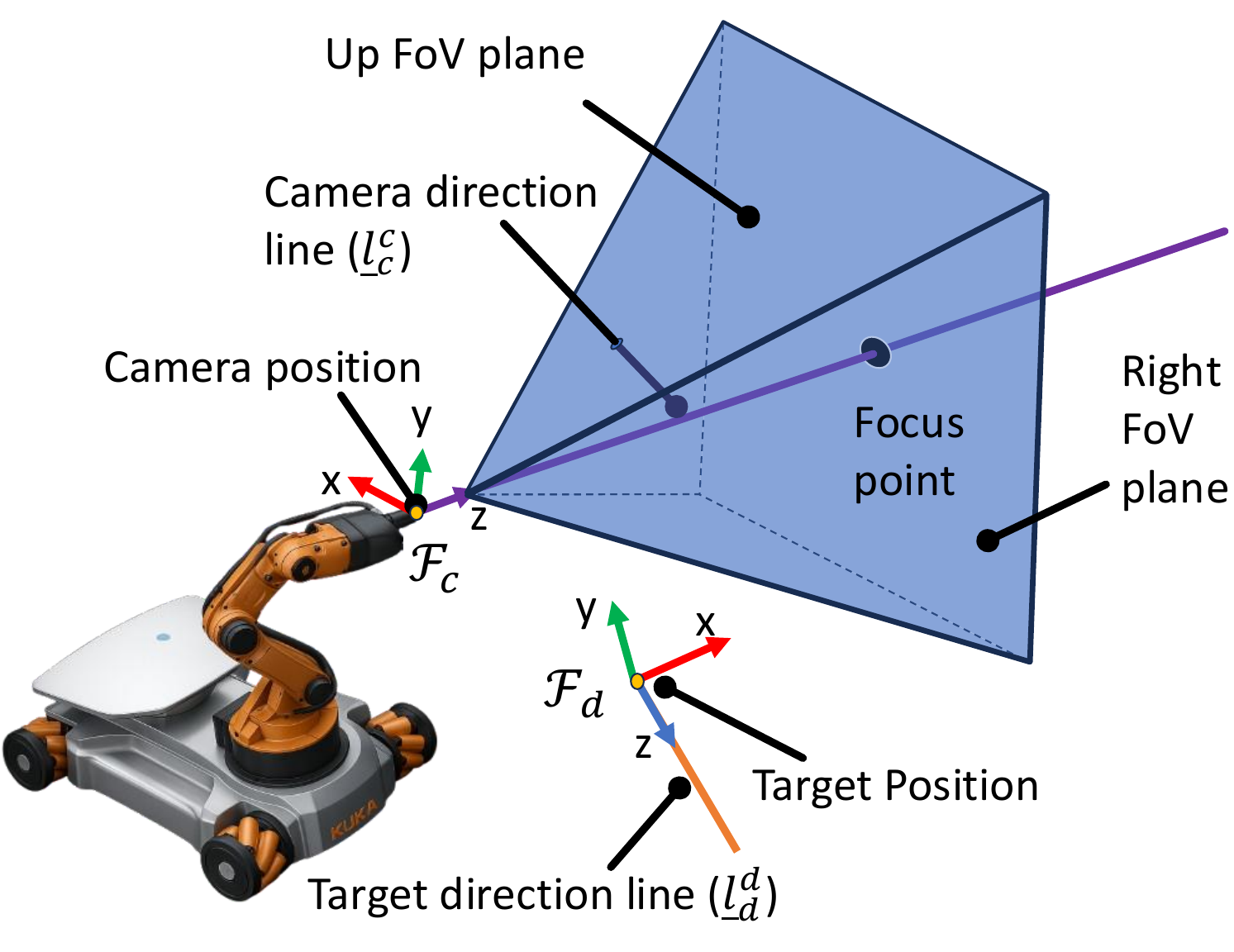}
      \caption{The illustration of the position and direction objectives.}
      \label{fig:position_direction}
\end{figure}

Without loss of generality, let us assume a camera is rigidly attached to the end-effector, such that their frames have the same origin and are aligned. Therefore, the camera position $\myvec{p}\in \mathbb{R}^3$ and direction $\myvec{l}\in \mathbb{R}^3$, with $\norm{\myvec{l}}_2=1$, vary according to the end-effector motion and depend on the robot configuration; namely, $\myvec{p} \triangleq \myvec{p}(\myvec{q}(t))$ and $\myvec{l} \triangleq \myvec{l}(\myvec{q}(t))$. 

Furthermore, let 
\begin{equation}
\myvec{x}\triangleq [\myvec{p}^T\quad \myvec{l}^T]^T \text{ and } \myvec{x}_d \triangleq [\myvec{p}_d^T\quad \myvec{l}_d^T]^T 
\end{equation}
be the current and  \emph{constant} desired task vectors, respectively, represented in the world frame.  To prescribe an exponential convergence of the end-effector position and direction to the desired set point, we define the error  $\error{\myvec{x}} \triangleq \myvec{x}-\myvec{x}_d$ with the \emph{desired} closed-loop error dynamics given by
\begin{equation}
\dot{\error{\myvec{x}}} + \lambda \error{\myvec{x}}=\myvec{0}\implies \mymatrix{J} \dot{\myvec{q}}+ \lambda \error{\myvec{x}} = \myvec{0},\label{eq:desired-error-dynamics}
\end{equation}
where $\mymatrix{J} = \left[\frac{\partial\myvec{p}}{\partial \myvec{q}}^T \quad \frac{\partial\myvec{l}}{\partial \myvec{q}}^T\right]^T$. The second equality in \eqref{eq:desired-error-dynamics} results from the fact that $\dot{\myvec{x}}_d=\myvec{0}$ for all $t$; hence, $\dot{\error{\myvec{x}}}=\dot{\myvec{x}}=\mymatrix{J}\dot{\myvec{q}}$. The first term in the objective function in \eqref{eq:control-law} is used to enforce the desired closed-loop error dynamics \eqref{eq:desired-error-dynamics} as best as possible.

\subsection{Soft Visibility Constraint and Objective Weighting}\label{sec:soft-visibility-constraint}
We simultaneously control the position and direction of the camera attached to the robot's end-effector. Therefore, some directions might conflict with the visibility constraint, creating challenges in satisfying both objectives concurrently, as illustrated in Fig.~\ref{fig:position_direction}. The visibility constraint requires the focus point to be within the set constrained by the four planes. However, aligning the camera direction line with the target direction line would violate the visibility constraint.

To overcome this problem, we have converted the hard visibility constraint from \cite{Ref:Dursun2023} into a soft one using the slack variable $\boldsymbol{\kappa}$ in \eqref{eq:control-law}. As the robot approaches the target position, we gradually reduce the importance of this constraint by using a distance-dependent weighting strategy similar to \cite{Ref:Holistic} and \cite{Ref:Visibility} because the robot is expected to prioritize the main objective of reaching the target position and direction, with visibility considered as a secondary objective. The weighting strategy is given by
\begin{equation}
    \lambda_\kappa=\alpha\norm{\error{\myvec{p}}}^\gamma_2,
\end{equation}
where $\norm{\error{\myvec{p}}}$ is the end-effector position error, and $\alpha,\gamma\in (0,\infty)$ determine how fast the weighting decays with this error \cite{Ref:Visibility}.

\subsection{Collision Avoidance and Circulation Constraints}
\label{subsec:collision_avoidance}
\subsubsection{Collision Avoidance Constraint for the End-Effector}\label{subsec:collision_avoidance_end_effector}
We represent the robot end-effector as a sphere with radius $r_\mathrm{eef}$, whereas the obstacles in the environment are modeled using boxes and cylinders. Collision avoidance is ensured by constraining the approach velocity between collidable entities using vector field inequalities (VFIs)\cite{Ref:VFI_Marinho}.

Let $\error{D}_\tau(\myvec{q})$, with $\tau \in \{\mathrm{box},~\mathrm{cyl}\}$, denote the signed distance to a box or a cylinder. Based on the signed distance function,  safe regions ($\omega_{S_i}$) and restricted regions ($\omega_{R_i}$) with respect to the $i$th obstacle are defined as  
\begin{align*}
    \omega_{R_i} \triangleq \{\myvec{q} \in \mathbb{R}^n :  \tau \in \{\mathrm{box},~\mathrm{cyl}\}, \error{D}_\tau(\myvec{q}) \leq 0\}, \\
    \omega_{S_i} \triangleq \{\myvec{q} \in \mathbb{R}^n :  \tau \in \{\mathrm{box},~\mathrm{cyl}\}, \error{D}_\tau(\myvec{q}) > 0\}.
\end{align*}
Accounting for $N_o$ obstacles, the global restricted and safe regions are given by
\begin{alignat*}{2}
\Omega_{R}\triangleq\bigcup_{i=1}^{N_o}\omega_{R_{i}} & \text{ and } & \Omega_{S}\triangleq\bigcup_{i=1}^{N_o}\omega_{S_{i}},
\end{alignat*}
such that $\Omega_R\cup\Omega_S=\mathcal{W}$, with $\mathcal{W}$  being the robot's workspace.

Consider a scalar distance function that encompasses the entire restricted zone $\Omega _R$. A natural way of defining this global distance function would be to consider the minimum of the individual distance functions to the $N_o$ obstacles, $D(\myvec{q})\triangleq\min\left( \error{D}_{\tau_1}(\myvec{q}), \error{D}_{\tau_2}(\myvec{q}), \ldots, \error{D}_{\tau_{N_o}} (\myvec{q})\right).$ Clearly, $D(\myvec{q})>0$ implies $\myvec{q}\notin\Omega_R$. However, the function $D$ is not differentiable everywhere because of the non-differentiability of the minimum function. To mitigate this problem, we use the $\mathrm{smin}$ function \cite{Ref:Circulation}: 
\begin{align}
\mathrm{smin}(\error{D}_{\tau_1}, \error{D}_{\tau_2}, \dots, \error{D}_{\tau_{N_o}}) \! \triangleq \!-h_m \ln\! \left( \!\frac{1}{N_o} \!\sum_{i=1}^{N_o} \negmedspace
\exp\!\left(\!-\frac{\error{D}_{\tau_i}}{h_m} \!\right)\!\right),
\label{eq:softmin}
\end{align}
where $h_m\in\left(0,\infty\right)$.

This function is always differentiable in its arguments (i.e., the $N_o$ distances to obstacles), as the distance functions themselves are also differentiable, and approximates the true minimum for small values of $h_m$. However, 
$\mathrm{smin}(\error{D}_{\tau_1}, \error{D}_{\tau_2}, \dots, \error{D}_{\tau_{N_o}}) \geq \min(\error{D}_{\tau_1}, \error{D}_{\tau_2}, \dots, \error{D}_{\tau_{N_o}})$ \cite{Ref:Circulation}, which means that the robot might consider itself farther from the closest obstacle than it actually is, potentially violating the real minimum distance. By defining $\hat{D}(\myvec{q})\triangleq  \mathrm{smin}(\error{D}_{\tau_1}, \error{D}_{\tau_2}, \dots, \error{D}_{\tau_{N_o}})$, one way to mitigate this issue is to subtract a margin $\delta>0$ from the estimated minimum;  that is, $\error{D}(\myvec{q}) =\hat{D}(\myvec{q})-\delta$ \cite{Ref:Circulation}.  The parameter $\delta$ is selected through trial and error. Larger values make the constraint more conservative \cite{Ref:Circulation}.

To calculate the gradient of $\error{D}(\myvec{q})$, let us define
\begin{align}
S(\myvec{q}) &\triangleq \frac{1}{N_o}\sum_{i=1}^{N_o} 
\exp\left(\!-\frac{\error{D}_{\tau_i}}{h_m} \right).
\end{align}
Using the derivative of the natural logarithm, we have
\begin{align}
\nabla \error{D}(\myvec{q}) = \nabla\hat{D}(\myvec{q})= -h_m\,\frac{1}{S(\myvec{q})}\,\nabla S(\myvec{q}),
\label{eq:gradient-of-D-smin}
\end{align}
where $\nabla S(\myvec{q})$ is given by
\begin{align}
\nabla S(\myvec{q})=-\frac{1}{N_oh_m}\sum_{i=1}^{N_o} 
\exp\left(-\frac{\error{D}_{\tau_i}}{h_m} \right) \nabla \error{D}_{\tau_i},
\label{eq:gradient-of-S-smin}
\end{align}
Plugging \eqref{eq:gradient-of-S-smin} back into \eqref{eq:gradient-of-D-smin}, we obtain
\begin{align}
\label{eq:smin_grad_D}
\nabla\error{D}(\myvec{q})
&= \sum_{i=1}^{N_o}
\frac{\exp\left(-\frac{\error{D}_{\tau_i}}{h_m} \right)}{\sum_{j=1}^{N_o} \exp\left(-\frac{\error{D}_{\tau_j}}{h_m} \right)}\nabla \error{D}_{\tau_i},
\end{align}
in which $\nabla \error{D}_{\tau_i}$ is the gradient of the signed distance to box or cylinder.

Because we want to ensure $\error{D}(\myvec{q}) \geq 0$ for all $t\geq 0$, it suffices to enforce the differential inequality $\dot{\error{D}}(\myvec{q}) + \lambda_d\error{D}(\myvec{q}) \geq 0$ at all times \cite{Ref:VFI_Marinho}. Thus, 
\begin{gather}
\nabla\error{D}(\myvec{q})^T \dot{\myvec{q}}+ \lambda_e\error{D}(\myvec{q}) \geq 0,
\label{eq:VFI-for-softmin}
\end{gather} 
which becomes the first inequality in \eqref{eq:control-law}.

To represent box and (finite) cylinder obstacles, we use the intersection of negative half-spaces defined by signed distance formulations for geometric primitives such as point-to-line and point-to-plane distances \cite{Ref:VFI_Marinho}. The maximum of the distances to the individual half-spaces provides a lower bound on the Euclidean distance for any point outside the intersection \cite{Ref:Hart_max}.

\begin{figure}[t]
      \centering
      \includegraphics[width=0.7\linewidth]{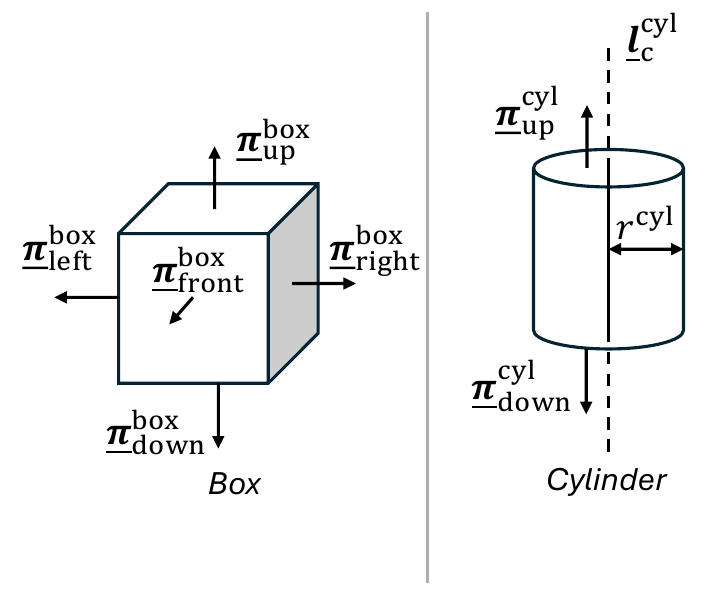}
      \caption{Box and cylinder obstacle boundaries are represented as compositions of infinite primitives.}
      \label{fig:softmax_obstacles}
\end{figure}

Fig.~\ref{fig:softmax_obstacles} demonstrate how the box-shaped collision boundaries are represented using six planes ($\dq{\pi}_\mathrm{up}^\mathrm{box}$, $\dq{\pi}_\mathrm{down}^\mathrm{box}$, $\dq{\pi}_\mathrm{left}^\mathrm{box}$, $\dq{\pi}_\mathrm{right}^\mathrm{box}$, $\dq{\pi}_\mathrm{front}^\mathrm{box}$, $\dq{\pi}_\mathrm{rear}^\mathrm{box}$), whereas a (finite) cylinder is represented using an infinite cylinder, which is defined by the centerline $\dq{l}_c^\mathrm{cyl}$ and radius $r^{\mathrm{cyl}}$, cut by two planes ($\dq{\pi}_\mathrm{up}^\mathrm{cyl}$,$\dq{\pi}_\mathrm{down}^\mathrm{cyl}$).

Considering a box shape, the signed distance between the sphere attached to the end-effector and the infinite planes can be written as $\error{d}_{p,\pi_\mathrm{j}^\mathrm{box}}\triangleq d_{p,\pi_\mathrm{j}^\mathrm{box}}-r_\mathrm{eef}$, where $j\in\{ \mathrm{up},~\mathrm{down},~\mathrm{left},~\mathrm{right},~\mathrm{front},~\mathrm{rear}\}$ and  $d_{p,\pi_\mathrm{j}^\mathrm{box}}$ is the signed distance between the  sphere center and the $j$th plane. Thus, a lower bound for the distance between the sphere attached to the end-effector and a box can be defined as $ D_\mathrm{box} (\myvec{q})\triangleq\max \mathcal{Y}_\mathrm{box}$ where $ \mathcal{Y}_\mathrm{box}\!=\! \{\error{d}_{p,\pi_\mathrm{j}^\mathrm{box}} \in \mathbb{R} : j \!\in\! \{\mathrm{up},\mathrm{down},\mathrm{left},\mathrm{right},\mathrm{front},\mathrm{rear}\}\}$.

 Similarly,  a lower bound for the distance between the sphere attached to the end-effector and a cylinder of radius $r_\mathrm{cyl}$ is defined as $D_\mathrm{cyl} (\myvec{q})\triangleq  \max \mathcal{Y}_\mathrm{cyl}$, where $\mathcal{Y}_\mathrm{cyl}= \{\error{d}_{p,\pi_\mathrm{up}^\mathrm{cyl}},~\error{d}_{p,\pi_\mathrm{down}^\mathrm{cyl}},~\error{d}_{p,l_\mathrm{c}^\mathrm{cyl}}\}$, with $\error{d}_{p,\pi_\mathrm{up}^\mathrm{cyl}}=d_{p,\pi_\mathrm{up}^\mathrm{cyl}} - r_\mathrm{eef}$ and  $\error{d}_{p,\pi_\mathrm{down}^\mathrm{cyl}} = {d}_{p,\pi_\mathrm{down}^\mathrm{cyl}}-r_\mathrm{eef}$ being the distance between the sphere and the top and bottom planes, and  $\error{d}_{p,l_\mathrm{c}^\mathrm{cyl}}=d_{p,l_\mathrm{c}^\mathrm{cyl}}-(r_\mathrm{eef}+r_\mathrm{cyl})$ is the distance between the sphere and the cylinder.

Although individual distance functions are differentiable, the functions $D_\mathrm{box}$ and $D_\mathrm{cyl}$ are not differentiable everywhere because of the non-differentiability of the maximum function, similar to the minimum function. Thus, a soft version of $\max$ function is used. Analogously to the property $\max(x_1,\ldots, x_n)\!=\!-\min(-x_1,\ldots,-x_n)$, direct calculation shows that $\mathrm{smax}(x_1,\ldots, x_n)\!=\!-\mathrm{smin}(-x_1,\ldots,-x_n)$. Therefore, the distance to the $i$th obstacle is approximated by $\error{D}_{\tau_i} = \mathrm{smax}~\mathcal{Y}_{\tau_i}+\delta$, where $\delta > 0$ is added because $\mathrm{smax}~\mathcal{Y}_{\tau_i} \leq \max~\mathcal{Y}_{\tau_i}$.

\subsubsection{Collision Avoidance Constraint for the Robot Base}

We use the same set of constraints for the robot base as for the robot end-effector, and the robot base is also modeled as a sphere. Thus, the aforementioned constraints, considering the robot base configuration, can be written as
\begin{gather}
\nabla\error{D}_b(\myvec{q}_\mathrm{base})^T \dot{\myvec{q}}_\mathrm{base}+ \lambda_d\error{D}_b(\myvec{q}_\mathrm{base}) \geq 0,
\label{eq:VFI-for-softmin_base}
\end{gather}
where $\error{D}_b$ is the $\mathrm{smin}$  distance function, given by \eqref{eq:softmin}, between a sphere enclosing the robot base and obstacles in the workspace, which is computed like the calculations for the end-effector presented in Section~\ref{subsec:collision_avoidance_end_effector}.

\subsubsection{Circulation Constraint for the Robot Base}
\label{subsubsec:circulation}
Although the constraints discussed in Section~\ref{subsec:collision_avoidance} effectively enable the robot to avoid collision with obstacles, the robot might get stuck in local minima, as the control law \eqref{eq:control-law} is convex but the overall problem is non-convex, making the closed-loop system only locally asymptotically stable. This typically happens when the robot is at the boundaries of obstacles, which can create a spurious equilibrium point. Gonçalves \emph{et al}. \cite{Ref:Circulation} have introduced a circulation constraint to mitigate the local minima problem by allowing the robot to circulate around the obstacles.

We adopt the circulation constraint only for the robot base to mitigate local minima. Their approach enables the robot to circulate around the obstacle by forcing the robot's velocities to align with the vector tangent to the distance gradient $\nabla\error{D}_b(\myvec{q}_\mathrm{base})$. The tangent vector to $\nabla\error{D}_b(\myvec{q}_\mathrm{base})$, denoted by $\mymatrix{T}(\nabla \error{D}_b)$, is such that $\left<\nabla\error{D}_b, \mymatrix{T}(\nabla \error{D}_b)\right> = 0$. One such vector can be defined as $ \mymatrix{T}(\nabla \error{D}_b) \triangleq \mymatrix{\Omega}\nabla \error{D}_b/||\mymatrix{\Omega} \nabla \error{D}_b||$, where $\mymatrix{\Omega} \in \mathbb{R}^{3 \times 3}$ is a skew-symmetric matrix given by

\begin{equation}
\label{eq:circ_mat}
\myvec{\Omega} =
\begin{bmatrix}
0 & 1 & 0 \\
-1 & 0 & 0 \\
0 & 0 & 0
\end{bmatrix},
\end{equation}
which forms a vector orthogonal to the vector $\nabla \error{D}_b$, as one can verify by direct calculation.

Lastly, a distance-dependent continuous function $\beta:\mathbb{R}\to\mathbb{R}$ is defined as
\begin{align}
    \beta(\error{D}_b) = b\left(1 - \frac{\error{D}_b}{D_0}\right),
\label{eq:betaeq}
\end{align}
where $b,D_0\in \left(0,\infty\right)$. This function is used with the tangent vector $\myvec{T}(\nabla\error{D}_b)$ to form the second inequality in \eqref{eq:control-law}. It ensures that if an obstacle is closer than an activation distance $D_0$ (i.e., $\error{D}_b<D_0$ ), then $\beta(\error{D}_b) > 0$ and hence $\myvec{T}(\nabla\error{D}_b)^T\dot{\myvec{q}}_\mathrm{base} \geq |\beta(\error{D}_b)|$, which activates circulation in the positive direction of $\myvec{T}(\nabla\error{D}_b)$, while any motion in the negative direction of $\myvec{T}(\nabla\error{D}_b)$ is forbidden. Conversely, when $\error{D}> D_0$ , $\beta(\error{D}_b) < 0$ and  $\myvec{T}(\nabla\error{D}_b)^T\dot{\myvec{q}}_\mathrm{base} \geq -|\beta(\error{D}_b)|$, which admits unbounded velocities in the positive direction of $\myvec{T}(\nabla\error{D}_b) $ and velocities bounded by $|\beta(\error{D}_b)|$ in the negative direction of $\myvec{T}(\nabla\error{D}_b)$. A greater $D_0$ means that circulation starts when the robot is farther from an obstacle. On the other hand, $b$ controls the circulation speed by tuning the effect of the constraint; it also influences how much the velocities in the negative direction of $\myvec{T}(\nabla\error{D}_b)$ are constrained when the robot is far from obstacles. For more details, please refer to \cite{Ref:Circulation}.

\section{Simulation and Experiments}{\label{experimental_setup}}

To demonstrate the performance of the proposed method, we conducted both simulations and real-world experiments.  The simulation setup enables us to simulate large workspaces, each containing a target object among a set of 114 large objects, providing meaningful data for statistical analysis. In contrast, the real-world experiments validate the method’s performance in practice on real robots. In simulations, we used a wheeled mobile manipulator consisting of a holonomic base and an arm with five joints. We employed two different mobile manipulators in the real-world experiments: one with a wheeled holonomic base and a five-joint arm, similar to the one used in the simulations, and a second robot with a six-joint arm mounted on a legged robot. The second robot was used to demonstrate the generality of the method across different classes of mobile manipulators.

\subsection{Simulation Setup}\label{sec:simulation-setup}
We performed simulations using CoppeliaSim \cite{Ref:coppeliaSim}, ROS Noetic \footnote{www.ros.org} and Python to validate our method. The simulations were performed on a computer with an Intel i9-12900H processor running at 2.9 GHz, 32 GB of RAM, and Ubuntu version 20.04. We used DQ~Robotics \cite{Ref:DQ} for rigid motion operations, robot modelling, description of geometric primitives, and the generation of control inputs given by \eqref{eq:control-law}. Furthermore, the PyRep library \cite{Ref:Pyrep} was used for communication between CoppeliaSim and Python.

To evaluate the coverage rate of the object, total entropy in the object bounding box, and the total time that the robot spends traveling between NBVs, we created 114 test environments in CoppeliaSim comprising a distinct large object from the ShapeNet \cite{Ref:Shapenet} dataset. 

\begin{figure}[t]
      \centering
      \includegraphics[width=1\linewidth]{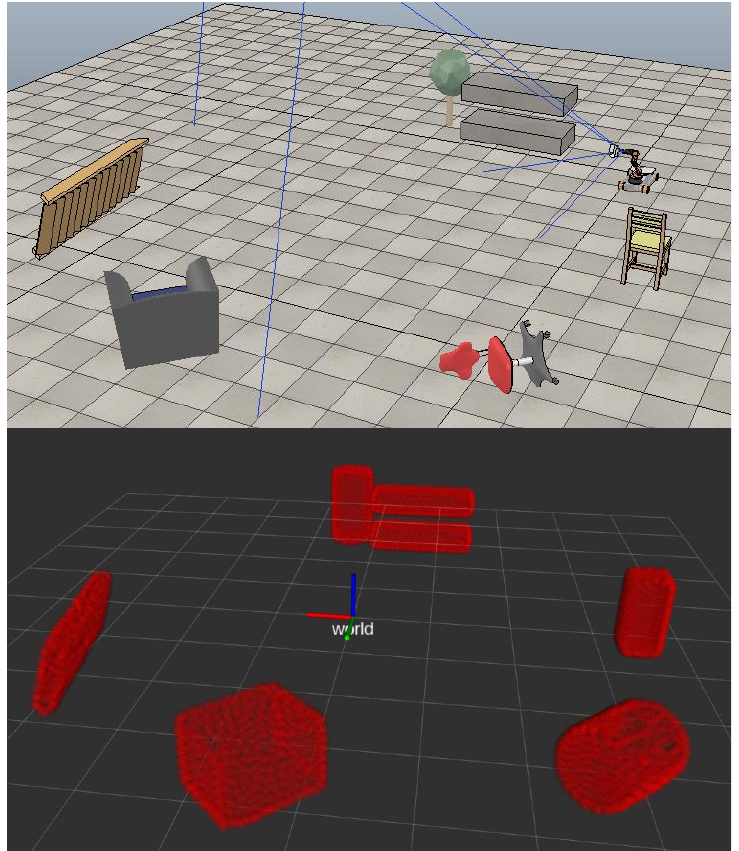}
      \caption{Illustration of the obstacles objects and their boxes and cylinders.}
      \label{fig:env_obstacles}
\end{figure}

Since the KUKA YouBot's base is holonomic and its arm has five joints, $n=8, n_b=3, n_a=5$ in \eqref{eq:control-law}.  Finally, we use four distance functions for four FoV planes to define the visibility constraint; hence, $n_v=4$. Table~\ref{tab:parameters} summarizes the parameters used in our implementation. Two situations were evaluated in a series of simulation runs. First, we collected reconstruction-related data in an environment without obstacles, except for a single virtual cylindrical forbidden region used to prevent the robot from entering the object area. Additionally, we demonstrate the collision avoidance performance of the proposed controller in a highly cluttered environment, where obstacles are modeled using boxes and cylinders, as shown in Fig.~\ref{fig:env_obstacles}.

\subsection{Comparison Metrics}
\subsubsection{Object Coverage} 
We compare the point cloud models acquired during the reconstruction process against the original model's point cloud to calculate the object coverage. The closest point in the reconstruction is found for each point in the original model. The surface point of the original model is considered as observed if it falls within a specified registration distance \cite{Ref:Isler}. The object coverage is given by \cite{Ref:Zeng} 
\begin{equation}
C(\mathcal{P})=\frac{1}{M_c}\sum_{\myvec{p}\in \mathcal{P}} U \left(\epsilon-\min_{\myvec{p}_0\in {\mathcal{P}_0\setminus \mathcal{P}^\prime_0}} \norm{\myvec{p}-\myvec{p}_{0}}\right),
\end{equation}
where $\mathcal{P}\subset \mathbb{R}^3$, with $|\mathcal{P}|=N_c$, is a partial point cloud with $N_c$ points, $\mathcal{P}_0\subset \mathbb{R}^3$, with $|\mathcal{P}_0|=M_c$, is the complete, reference point cloud of the object containing $M_c$ points, $\mathcal{P}^\prime_0=\{\myvec{p}^\prime_0\in P_0 : \myvec{p}^\prime_0~\text{is covered}\}$ is the set of points already covered in the reference point cloud, $U$ is the unit step function, and $\epsilon=8~\mathrm{mm}$ is the registration distance.

We also report the Area-Under-the-Curve (AUC) metric \cite{Ref:Delmerico}, which illustrates the object coverage speed of the different methods. This metric is included because, although the methods may reach similar final coverage values, some methods achieve the same coverage at earlier NBV stages.

\subsubsection{Entropy}
To compute the total entropy, we define a bounding box with a width and length of 5.5~m and a height of 4~m around the object and add the entropy of all voxels inside of it:
\begin{equation}
    E=\sum_{x\in \mathcal{U}} \mathcal{I}(x),
\end{equation}
where $\mathcal{U}$ is the set of all the voxels within the bounding box, and $\mathcal{I}(x)$ is the voxel entropy given by \eqref{eq:voxel_entropy}.

\subsubsection{Total Time ($T_\mathrm{total}$)}
We compare the total time that the robot spends traveling between all NBVs. The total time is composed of two components:
\begin{enumerate}
    \item Computation time ($T_\mathrm{comp}$): It is the total time spent calculating informative path between NBVs. This represents the total time required to sample and evaluate candidate views along the paths for the sampling-based method, and the total time required to compute all focus points for our method until visiting all NBVs. This metric corresponds to the sum of $t_\mathrm{eval}$ in Fig.~\ref{fig:ipp_exec_eval} across all NBVs; that is, $T_\mathrm{comp} = \sum_{i=1}^{N_\mathrm{nbv}} t_{\mathrm{eval}_i}$, where $N_\mathrm{nbv}$ is the number of NBVs.
    \item Execution time ($T_\mathrm{exec}$): The total robot motion time to visit all NBVs, excluding all computations between NBVs. This metric corresponds to the sum of $t_\mathrm{exec}$ in Fig.~\ref{fig:ipp_exec_eval} across all NBVs; that is, $T_\mathrm{exec}=\sum_{i=1}^{N_\mathrm{nbv}} t_{\mathrm{exec}_i}$.
\end{enumerate}
It should be noted that the summary statistics provided in the tables are the average values of these metrics over the total number of objects.

For all methods, we used the same controller defined in \eqref{eq:control-law} without changing its parameters. The only difference is that our proposed approach includes a visibility constraint to maintain a focus point within the camera's field of view, whereas the other two methods do not include any visibility constraints. Therefore, as the motion controller is the same for all strategies, the measured time differences are not affected by the controller and depend solely on the strategies used on the way to the NBVs to enhance object reconstruction performance.
\subsection{Data Collection}
We employed the rear-side-voxel method to determine the NBVs that are selected among 200 discrete candidate views generated around a vertical cylindrical search space, as explained in Section~\ref{sec:focus_point_calculation}. After calculating the NBVs, we considered our strategy, denoted \textbf{focus}, and two other baseline strategies (\textbf{no path} and \textbf{sampling}) to determine the robot's behaviour between two consecutive NBVs: 
\begin{enumerate}
\item \textbf{Focus}: the robot keeps an informative focus point using the strategy proposed in Section~\ref{sec:focus_point_calculation}.  In both simulation and experiments, a new focus point is computed whenever the robot’s end-effector moves more than \unit[1]{m} from the position where the previous focus point was calculated.
\item \textbf{No path}: the robot moves toward the NBV without focusing on any particular point; that is, (\ref{eq:control-law}) is used without a visibility constraint to control the robot.
\item \textbf{Sampling}: the robot uses an RRT* sampling-based method to plan an informative path between NBVs. The method relies on using nodes of the RRT* to sample and evaluate views using ray-casting operations along the path between NBVs, representing the current state-of-the-art IPP methods \cite{Ref:Song_1,Ref:Song_2,Ref:Bircher}. In addition, the ray-casting process is performed using the same parameters as in our focus point estimation method to ensure a fair comparison in terms of view evaluation. The details can be found in the Appendix\ref{sec:appendix}. 
\end{enumerate}

For each object, we collected data for 10 NBVs regarding the object coverage, entropy at the final NBV, and the total time the robot spent visiting all 10 NBVs.
\begin{figure*}[t]
  \centering
  \begin{subfigure}[b]{0.67\columnwidth}
    \includegraphics[width=\linewidth]{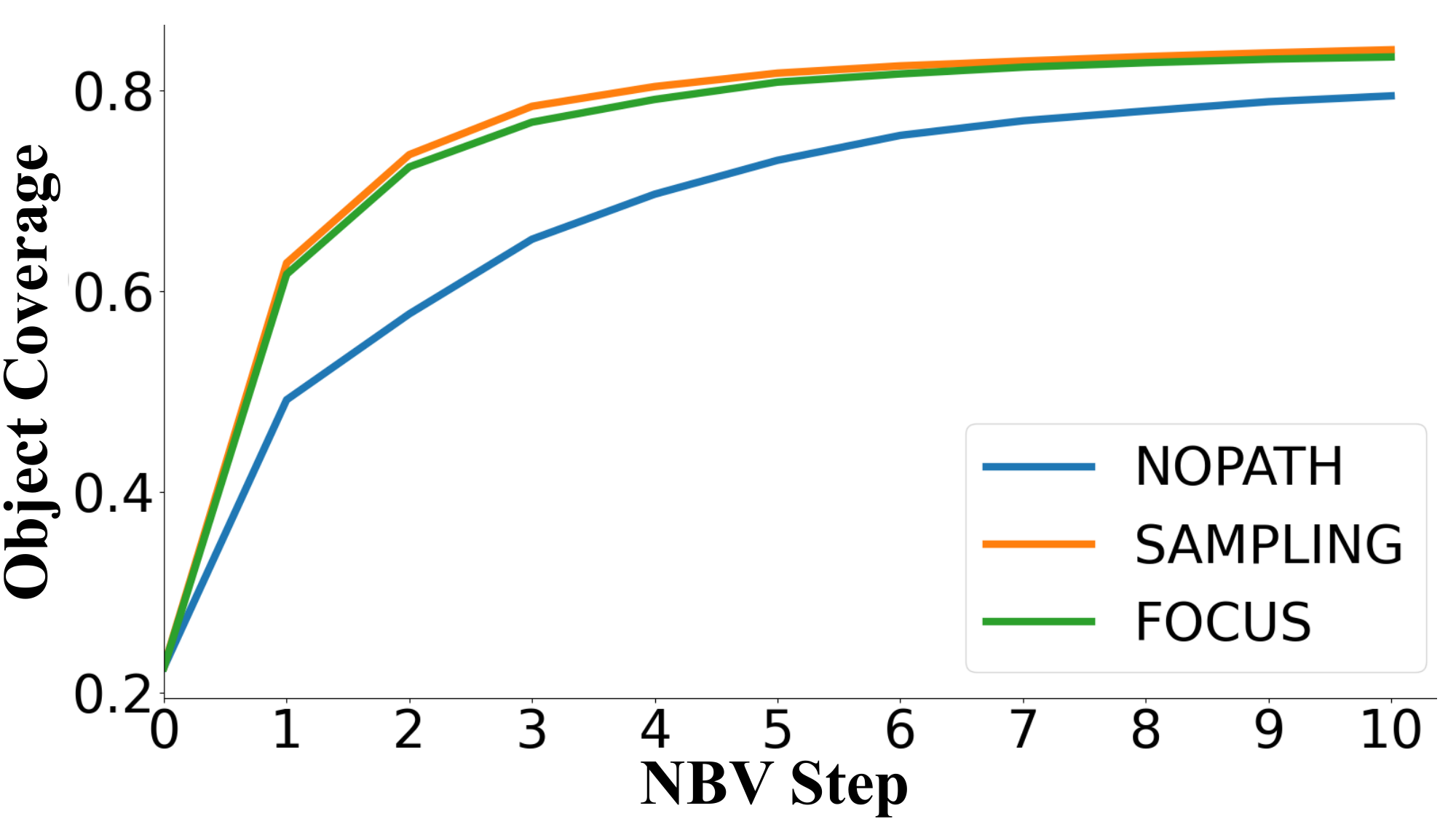}
    \caption{}
   
  \end{subfigure}
  \begin{subfigure}[b]{0.67\columnwidth}
    \includegraphics[width=\linewidth]{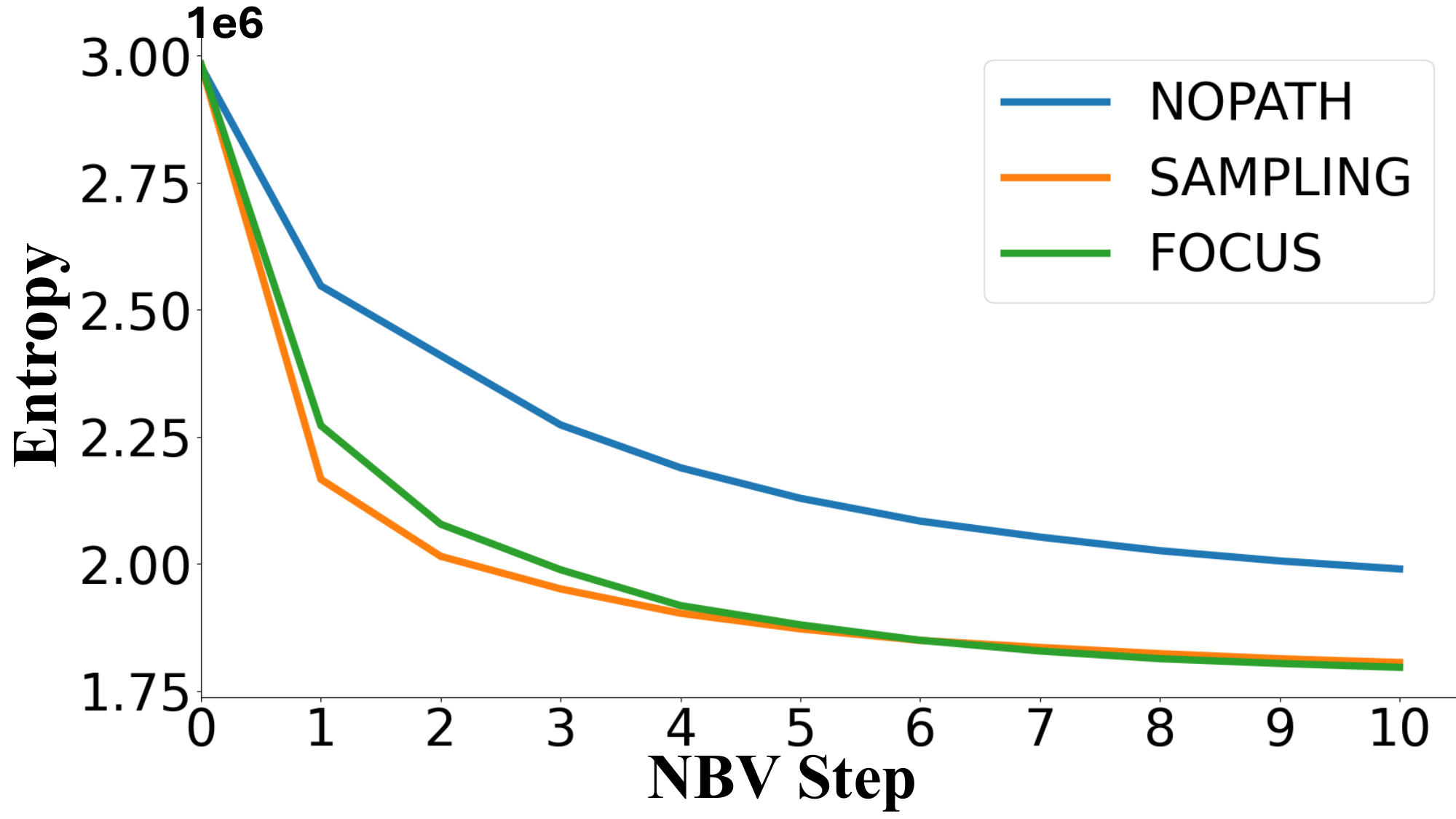}
    \caption{}
  
  \end{subfigure}
  \begin{subfigure}[b]{0.67\columnwidth}
    \includegraphics[width=\linewidth]{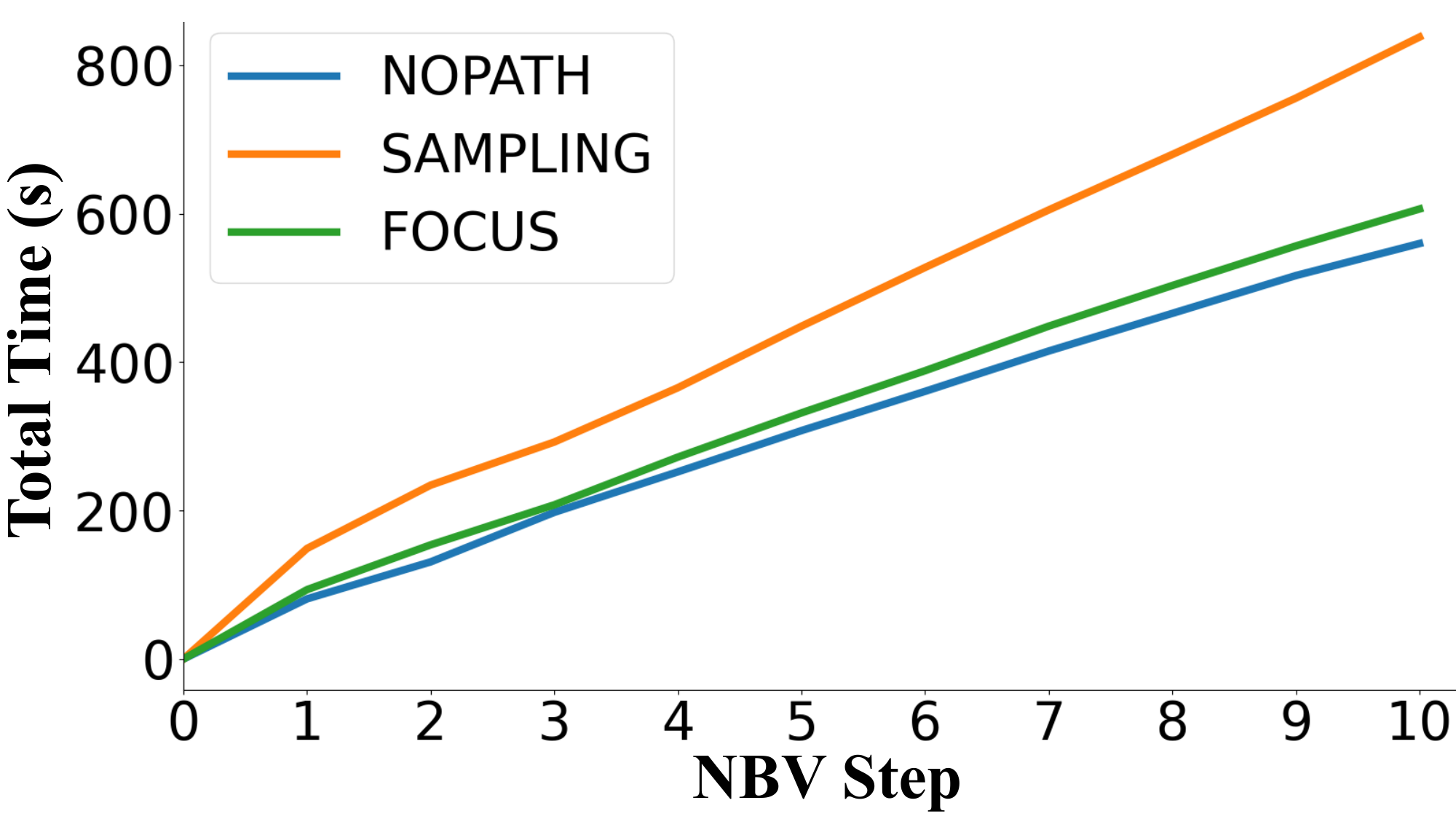}
    \caption{}
   
  \end{subfigure}
  \caption{Average object coverage, entropy, and total time for consecutive NBV steps across 114 runs in simulation. Our method (\textbf{focus}) and the \textbf{sampling} method both demonstrate similar object coverage and entropy performance, consistently outperforming the \textbf{no path} method. Additionally, our method requires significantly less time than the \textbf{sampling} method.}
  \label{fig:cov_entropy_time_avg}
\end{figure*}
\begin{table}[t]
\caption{Parameters used in our implementation}

\label{tab:parameters}
\begin{center}
\begin{tabular}{ccl}
Parameter & Value & Description\\
\hline
\hline
$r$ & 0.03~m & model resolution\\
\hline
$d_{\mathrm{max}}^{\mathrm{sensor}}$ & 4.5~m & maximum distance of depth sensor\\
\hline
FoV&($74^{\circ}$, $60^{\circ}$)& \makecell[l]{horizontal and vertical angles of the \\camera FoV}\\
\hline
\makecell[l]{ Expanded\\FoV} &($90^{\circ}$, $90^{\circ}$)& \makecell[l]{ horizontal and vertical angles of the FoV used \\for the focus point calculation}\\
\hline
$d_f$&2.5~m&\makecell[l]{distance between the focus point\\ and the camera center}\\
\hline
$d_{\mathrm{th}}$&0.75~m&\makecell[l]{the threshold distance for \\the visibility constraint}\\
\hline
$\lambda$&5& controller gain\\
\hline
$\lambda_q$&0.01& damping factor\\
\hline
$b$ &0.044~m/s& circulation speed factor\\
\hline
$D_0$ & 0.14~m& activation distance for circulation constraint \\
\hline
$\delta$ &0.05~m& offset distance in the $\mathrm{smin}$ function \\
\hline
$h_m$&0.03& $\mathrm{smin}$ parameter\\
\hline
\end{tabular}
\end{center}
\end{table}

\begingroup
\setlength{\tabcolsep}{4pt} 

\begin{table}[b]
\caption{The mean ($\mu$) and standard deviation ($\sigma$) of object coverage, AUC, entropy, and total time ($T_\mathrm{total}=T_\mathrm{comp}+T_\mathrm{exec}$) along with computation time ($T_\mathrm{comp}$) and execution time ($T_\mathrm{exec}$) for the \textbf{focus}, \textbf{sampling} and \textbf{no~path} methods at the 10\textsuperscript{th} NBV.}
\label{tab:cov_ent_time_stats}
\begin{center}

\begin{tabular}{m{1.9cm}m{1.95cm}m{1.95cm}m{1.95cm}}
 &Focus&Sampling&No~path\\
\hline
\hline
&$\mu\ (\sigma)$&$\mu\ (\sigma)$&$\mu\ (\sigma)$\\
Coverage&0.833 (\underline{\textbf{0.093}})&\underline{\textbf{0.84}} (0.094)&0.794 (0.103)\\
AUC&0.684 (\underline{\textbf{0.083}})&\underline{\textbf{0.693}} (0.086)&0.613 (0.103)\\
Entropy ($\times 10^3$)&\underline{\textbf{1796.25}} (128.95)&1805.75 (\underline{\textbf{123.30}})&1989.85 (155.32)\\
\hline
$T_\mathrm{comp}$ (s) &\underline{\textbf{11.7}} (\underline{\textbf{3.24}})&203.61 (74.03)&\quad\textemdash\\
$T_\mathrm{exec}$ (s)&588.76 (124.43)&629.81 (136.74)&\underline{\textbf{555.37}} (\underline{\textbf{81.65}})\\
$T_\mathrm{total}$ (s)&600.46 (126.72)&833.42(205.05)&\underline{\textbf{555.37}} (\underline{\textbf{81.65}})\\
\hline
\end{tabular}

\end{center}
\end{table}
\endgroup

\subsection{Analysis}
\subsubsection{Summary Statistics}
 The average object coverage, entropy, and total time calculated using 114 objects at each NBV step for 10 NBVs are illustrated in Fig.~\ref{fig:cov_entropy_time_avg}. Our method (\textbf{focus}) and the \textbf{sampling} method consistently improve object coverage and entropy compared to \textbf{no path}. Furthermore, our method requires significantly less time than the \textbf{sampling} method. 
 
The mean ($\mu$) and standard deviation ($\sigma$) values at the 10\textsuperscript{th} NBV step for three strategies are provided in Table~\ref{tab:cov_ent_time_stats}. The results show that our method reaches a slightly lower mean object coverage value of 0.833 than the \textbf{sampling} method (0.84) with similar standard deviations (0.093 and 0.094, respectively). Furthermore, the \textbf{no~path} results in a coverage value of 0.794, which is around 4\% less than the \textbf{focus} and \textbf{sampling}. Similar to the coverage results, \textbf{focus} and \textbf{sampling} show comparable AUC values of 0.684 and 0.693, respectively, which are significantly higher than \textbf{no~path}, which has an AUC of 0.613. This indicates that focus and sampling not only achieve better final coverage but also reach it faster than the \textbf{no~path} method. Considering entropy, our method resulted in an entropy of $1796.25\cdot 10^3$, indicating that it performs very similar to the \textbf{sampling} strategy ($1805.75\cdot 10^3$). Furthermore, both outperform the \textbf{no~path} strategy, which resulted in an  entropy of $1989.85\cdot 10^3$. Considering the total time ($T_\mathrm{total}$), our method (\textbf{focus}) requires 600.46 seconds across all NBVs, which is approximately 8.12\% more than \textbf{no~path} (555.37~s), while the \textbf{sampling} method requires 833.42 seconds, approximately 50.06\% more than \textbf{no~path}. Compared directly with \textbf{sampling} (833.42~s), our method (600.46~s) is 27.9\% faster in terms of total elapsed time ($T_\mathrm{total}$) across all NBVs (i.e., ($833.42-600.46)/833.42$).

Table~\ref{tab:cov_ent_time_stats} illustrates the time components used to calculate the total time metric for all methods. The greatest  time difference is the computation time ($T_\mathrm{comp}$), which is 11.7 seconds for our method and 203.61 seconds for the sampling method. This renders our method approximately 17.4 times faster than the \textbf{sampling} method with respect to computation time. This result is expected since the sampling-based method sample and evaluate many views on the path to the NBV, whereas our method only calculates focus points with an expanded FoV. This aligns with the computational complexity analysis. In this setting, the total number of evaluated view samples, averaged over the number of objects, is 1697.6; that is, $KN_\mathrm{sampling} = 1697.6$. The average number of calculated focus points is 52.75; that is, $N_\mathrm{focus} = 52.75$. As shown in Section \ref{subsec:focus_point_method},  $a_ha_v = 1.83$ when  $F_s = (74^\circ,60^\circ)$ and $F_f = (90^\circ, 90^\circ)$. Therefore, the computational complexity ratio between the two methods, as determined by \eqref{eq:complexity_relation}, is given by $1697.6 / (1.83 \cdot 52.75) = 17.58$, which is close to 17.4.

Moreover, our method spends 588.76 seconds executing the motion ($T_\mathrm{exec}$), while the sampling method requires 629.81 seconds, which is 6.97\% longer than our method. This result is also expected since sampling method requires the robot to visit many intermediate views, requiring convergence to each view, whereas our method continuously directs the camera toward informative regions without the need to stop between NBVs.

\subsubsection{Bayesian Data Analysis}
 
Although Fig.~\ref{fig:cov_entropy_time_avg} and the summary statistics presented in Table~\ref{tab:cov_ent_time_stats} enable a rough interpretation of method performance, they fail in addressing the question of whether the difference between methods is statistically significant and to what extent. This limitation becomes evident when examining the comparative results presented in the literature \cite{Ref:Isler,Ref:Delmerico,Ref:Daudelin}, as it is challenging to conclude which method is superior from a statistical standpoint. The limitation arises from the fact that focusing solely on mean and standard deviation fails to offer a comprehensive performance comparison across the general distribution of methods' performance for various objects. To this end, we conduct further analysis using Bayesian data analysis. 

Bayesian data analysis is a statistical approach that uses the Bayesian rule to make inferences about unknown parameters in a statistical model. Bayesian analysis treats parameters as random variables with prior beliefs that are updated when new data become available. \cite{Ref:Bayesian}. In our case, the parameters are the comparison metrics, namely object coverage, entropy, and time that the robot spends between NBVs.

Since the object coverage values for the \textbf{focus} and \textbf{sampling} methods reach steady-state at the 10\textsuperscript{th} NBV step (Fig.~\ref{fig:cov_entropy_time_avg}), we employed data belonging to this stage to perform Bayesian data analysis.  We analyzed the difference of means using the highest density interval (HDI) and the region of practical equivalence (ROPE). The HDI represents the interval of values more credible than those outside of it. For example, a 95\% HDI means that the interval has a total probability density of 95\%. The ROPE indicates the range of parameter values considered practically equivalent to the null value for a particular application. If the 95\% HDI of the difference of means between two distributions is completely inside the ROPE, the parameters causing these distributions are considered practically equivalent. Conversely, if the 95\%  HDI falls outside the ROPE, the parameters causing these distributions are practically nonequivalent. If the 95\%  HDI and the ROPE partially overlap, further analysis should be conducted \cite{Ref:Bayesian}.

We use large and diverse objects, and the candidate view space may not reveal some objects completely. Thus, some parameter values (object coverage, entropy, and time) could diverge significantly from the mean, thereby becoming outliers. Hence, we employed the $t$ distribution as the likelihood function for the difference of means, which ensures robustness against these outliers by adjusting the heaviness of its tails \cite{Ref:Bayesian}. We chose a prior normal distribution with a high standard deviation for the mean ($\mu$) of the $t$ distribution to avoid introducing strong prior information. Since we do not have prior information on the scale parameter ($\sigma$) of the $t$ distribution, we chose a prior uniform distribution. Lastly, we used an exponential distribution for the normality parameter ($\nu$ ) of the $t$ distribution to ensure significant credibility for low values, as the $t$ distribution varies mostly when $\nu$ is relatively small \cite{Ref:Bayesian}.

\subsubsection{Difference of Means Analysis}
\begin{figure*}[t!]
  \centering
  \begin{subfigure}[b]{0.3\textwidth}
    \includegraphics[width=\linewidth]{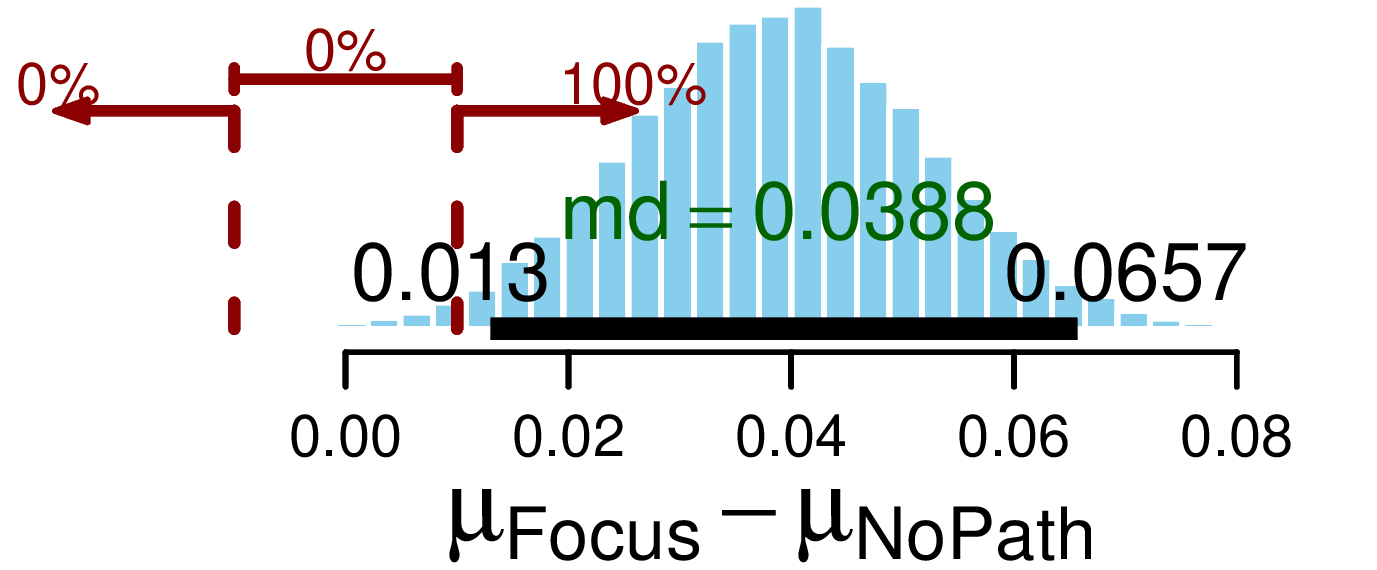}
    \caption{}
  \end{subfigure}
  \qquad
  \begin{subfigure}[b]{0.3\textwidth}
    \includegraphics[width=\linewidth]{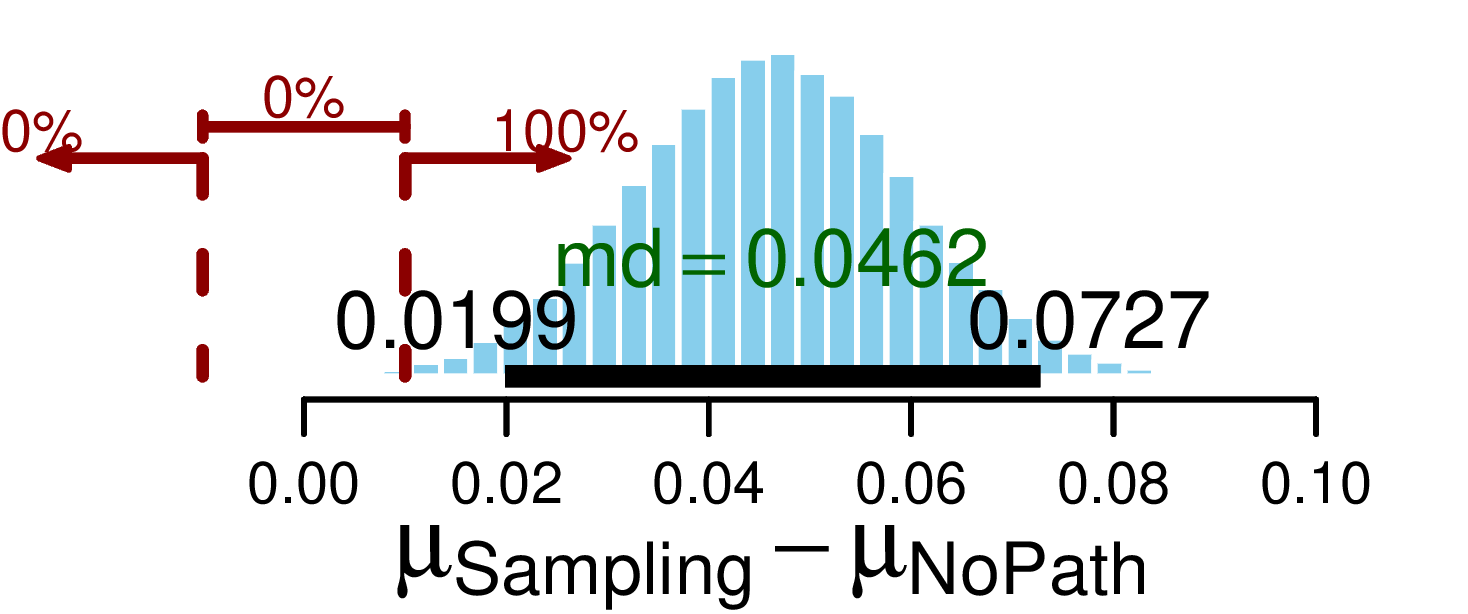}
    \caption{}
  \end{subfigure}
  \qquad
  \begin{subfigure}[b]{0.3\textwidth}
        \includegraphics[width=\linewidth]{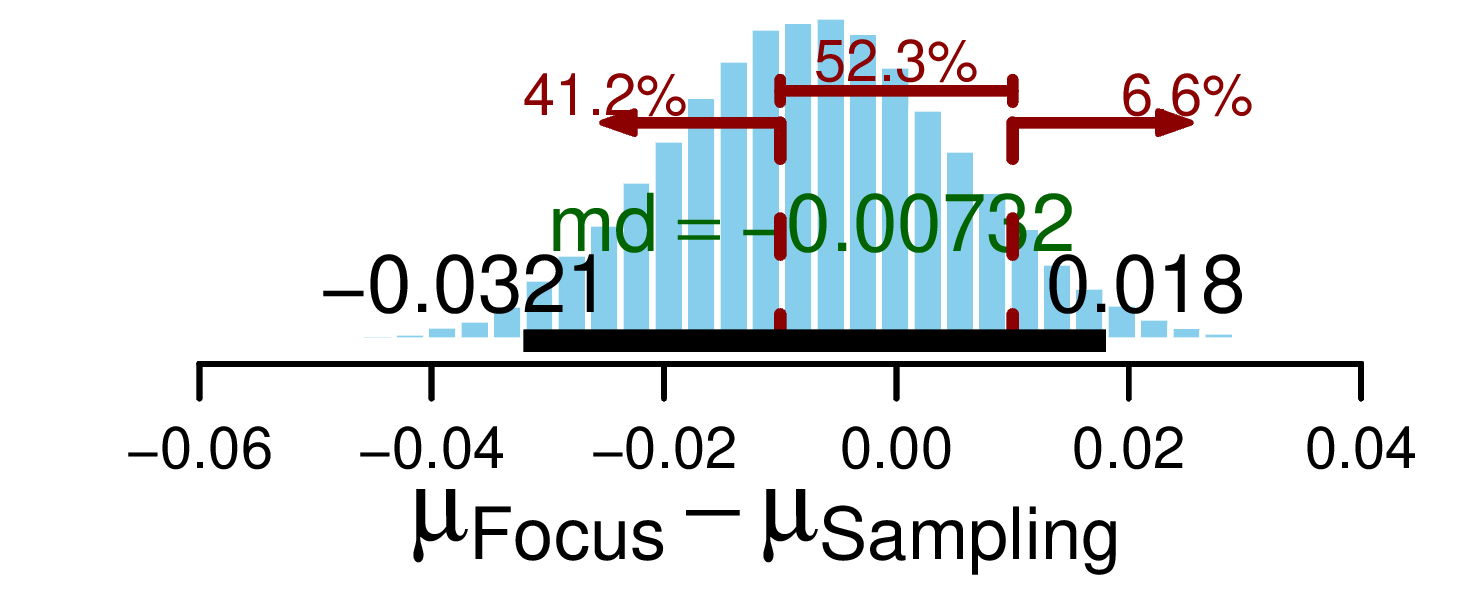}
    \caption{}
  \end{subfigure}
  
  \begin{subfigure}[b]{0.3\textwidth}
    \includegraphics[width=\linewidth]{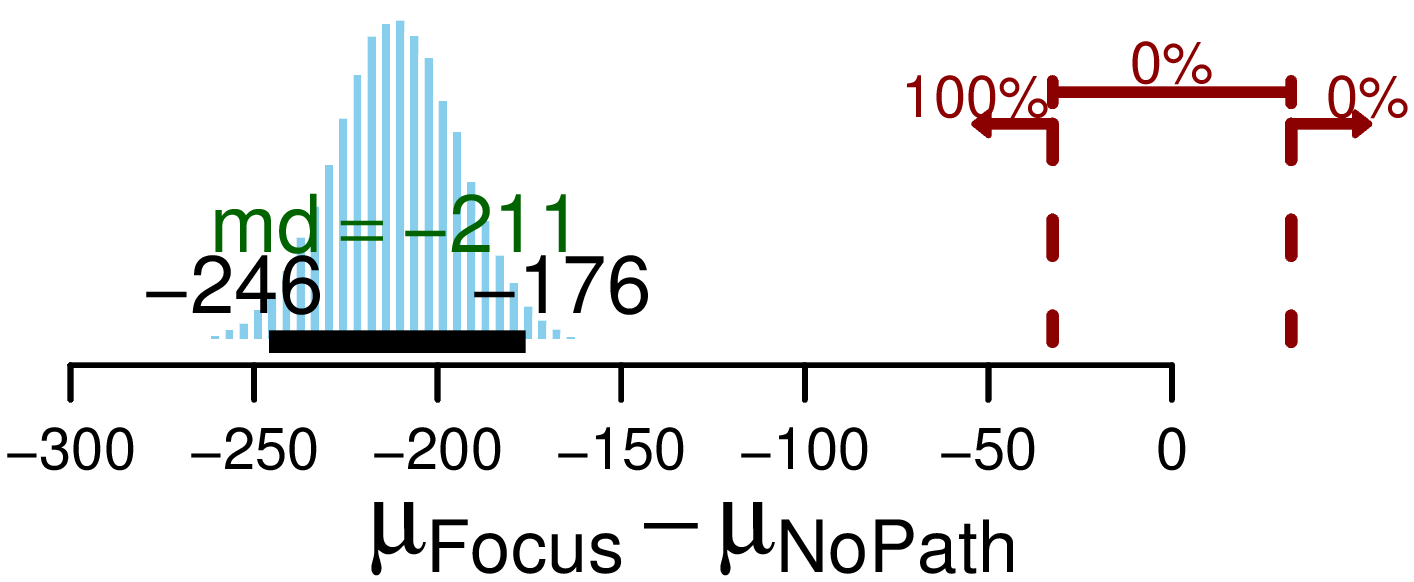}
    \caption{}
  \end{subfigure}
  \qquad
  \begin{subfigure}[b]{0.3\textwidth}
    \includegraphics[width=\linewidth]{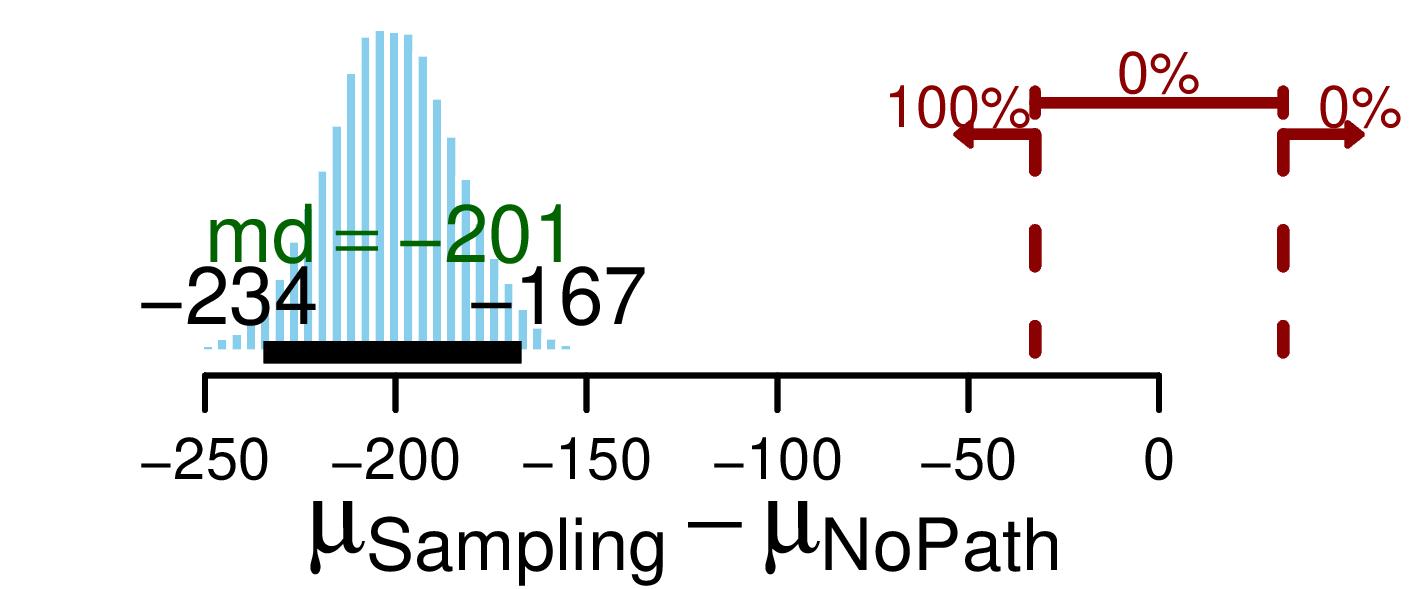}
    \caption{}
  \end{subfigure}
  \qquad
  \begin{subfigure}[b]{0.3\textwidth}
    \includegraphics[width=\linewidth]{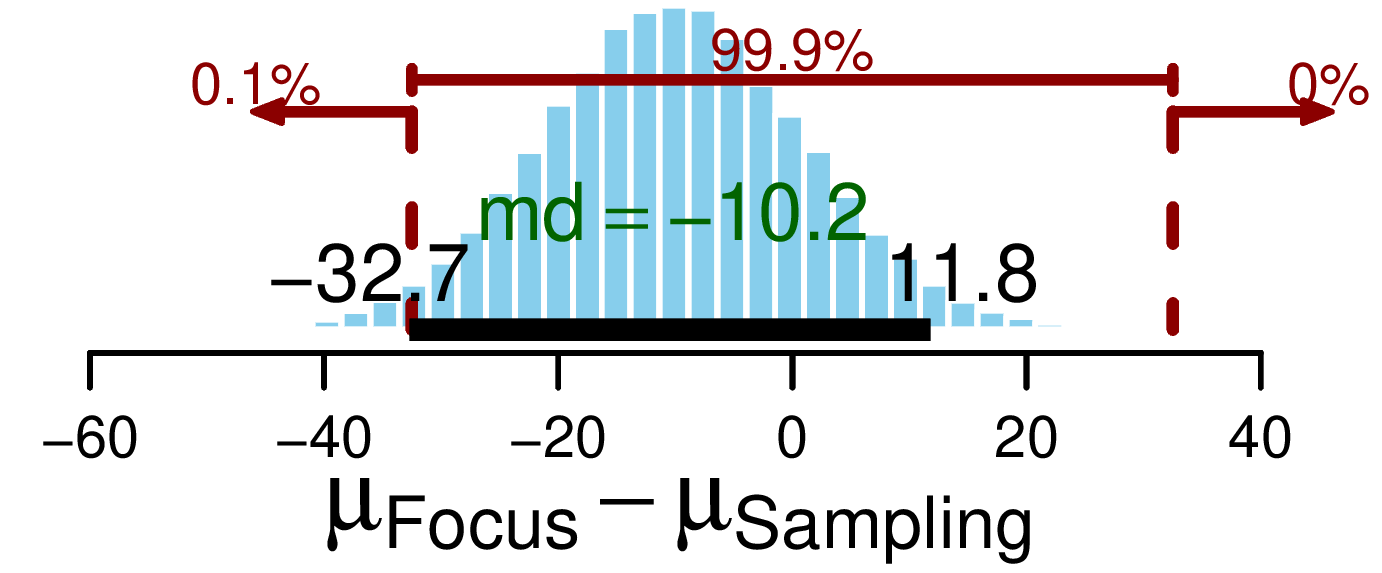}
    \caption{}
  \end{subfigure}

  \begin{subfigure}[b]{0.30\textwidth}
    \includegraphics[width=\linewidth]{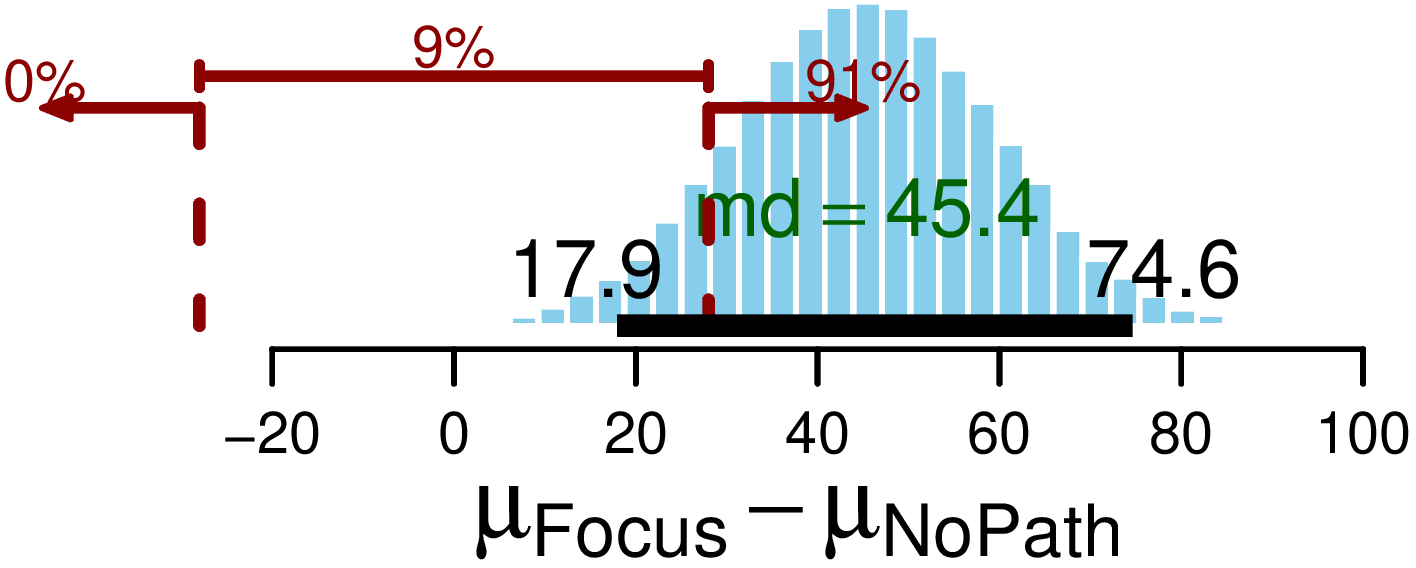}
    \caption{}
  \end{subfigure}
  \qquad
    \begin{subfigure}[b]{0.30\textwidth}
    \includegraphics[width=\linewidth]{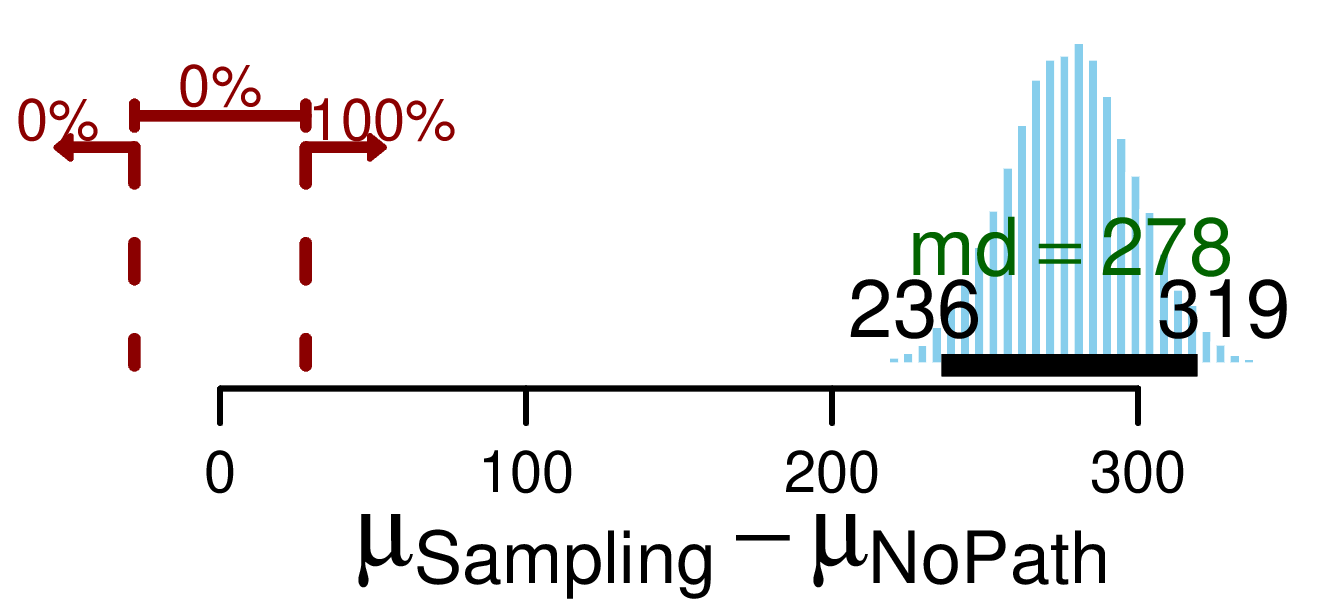}
    \caption{}
  \end{subfigure}
  \qquad
  \begin{subfigure}[b]{0.30\textwidth}
    \includegraphics[width=\linewidth]{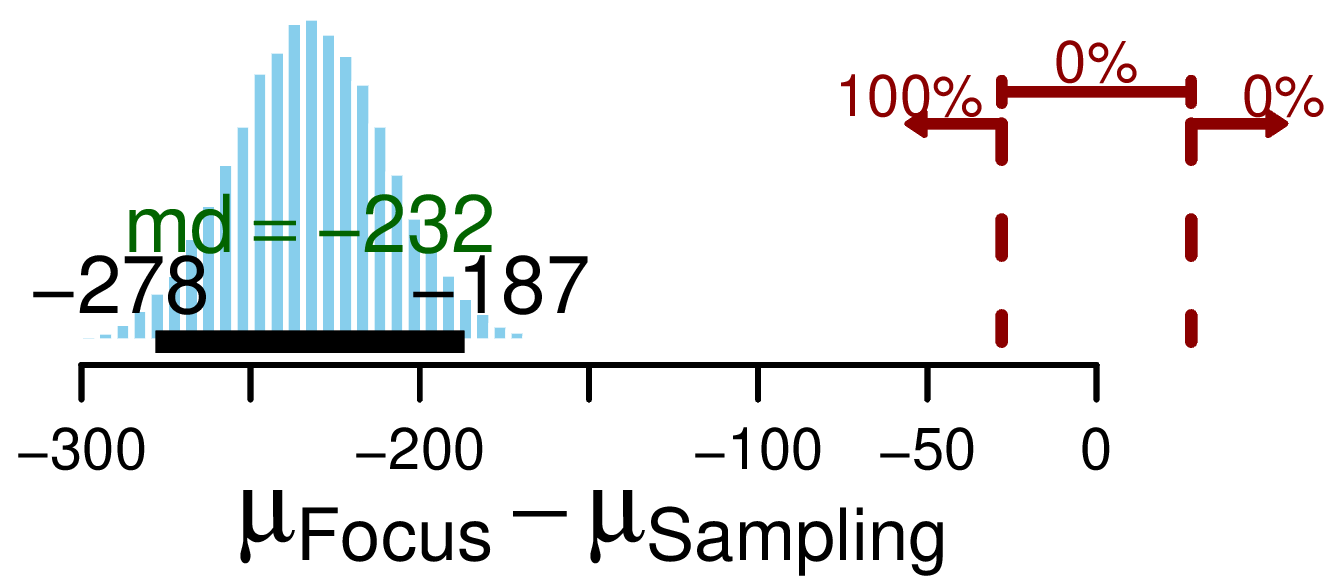}
    \caption{}
  \end{subfigure}
  \caption{The difference of means from the simulation data between the three methods, \textbf{focus} ($\mu_{\mathrm{focus}}$), \textbf{sampling} ($\mu_{\mathrm{sampling}}$), and \textbf{no~path} ($\mu_{\mathrm{NoPath}}$), for the object coverage (\emph{a--c}), entropy (\emph{d--f}) and total time, $T_\mathrm{total}$ (\emph{g--i}). Our method (\textbf{focus}) and the \textbf{sampling} method show similar performance regarding object coverage and entropy, while both significantly outperform the \textbf{no~path} method. Additionally, our method significantly reduces computation time compared to the \textbf{sampling} method.}
  \label{fig:difference_means}
\end{figure*}

We use the difference of means to compare the performance of the three strategies, namely \textbf{focus}, \textbf{no~path}, and \textbf{sampling}, shown in Fig.~\ref{fig:difference_means}. The red dashed lines represent the ROPE interval, and the black horizontal line represents the 95\% HDI. According to the literature \cite{Ref:Isler,Ref:Delmerico,Ref:Daudelin}, a 1\% difference in object coverage could be considered significant; therefore, we determine the ROPE limits as $[-0.01, 0.01]$ for object coverage. We consider a negative difference (less entropy is better) in at least 1\% of the maximum possible entropies in the object bounding box as a significant improvement.\footnote{The literature does not present any guidance on how to precisely quantify the improvement based on the entropy. Therefore, although we have chosen this threshold arbitrarily, further studies might improve this definition as the scientific community starts conducting more rigorous statistical analyzes.} Due to the large magnitudes of entropy values (Fig.~\ref{fig:cov_entropy_time_avg}b), we divide them by 1000 for analysis, as this does not affect the comparison results. This adjustment results in a maximum entropy of around 3250, where 1\% of this value is 32.5. Thus, we use $[-32.5, 32.5]$ as the ROPE interval for the entropy. Finally, to determine the ROPE interval for the time metric, we use the average time the robot spends between NBVs with the \textbf{no~path} method, 555.37  seconds. We consider any difference of less than 5\% of this time to be equivalent, implying a ROPE of $[-28, 28]$.

\paragraph{Object Coverage}
Fig.~\ref{fig:difference_means}(\emph{a--c}) shows the difference of means between the three methods for the object coverage. Fig.~\ref{fig:difference_means}(\emph{a}) shows that the ROPE and HDI do not overlap when comparing our method (\textbf{focus}) with the \textbf{no~path} baseline, meaning that those two methods are statistically distinct. Moreover, since the difference of means ($\mu_\mathrm{focus}-\mu_\mathrm{NoPath}$) is positive, with a median of 0.0388, the coverage using our method is 3.88\% higher than when not selecting intermediate views along the path between two consecutive NBVs. The \textbf{sampling} method also outperforms the \textbf{no~path} baseline with a median of 4.62\% of improvement. Finally, when comparing the \textbf{focus} and \textbf{sampling} strategies, the ROPE is completely inside the HDI. In that case, we cannot definitively say whether they are equivalent or non-equivalent using current data and the current ROPE. 

\paragraph{Entropy}

Fig.~\ref{fig:difference_means}(\emph{d--f}) shows the difference of means between the three methods for the entropy. The entropy performance of the \textbf{focus} strategy is statistically distinct from the \textbf{no~path} strategy because the ROPE does not overlap with the HDI. Moreover, the \textbf{focus} strategy yields smaller entropy than the \textbf{no~path} baseline, with a median difference of -211. This means an improvement of 6.5\%  over the maximum possible entropy (i.e., $-211/3250$). Considering the comparison between the \textbf{sampling} and \textbf{no~path} strategies, the ROPE does not overlap with the HDI, and \textbf{sampling} improves the entropy performance by 6.18\% over the maximum possible entropy (i.e., $-201/3250$) compared to \textbf{no~path}. Finally, the \textbf{focus} and \textbf{sampling} methods are statistically equivalent in terms of entropy since the entire HDI lies within the ROPE limits.

\paragraph{Total Time}
Fig.~\ref{fig:difference_means}(\emph{g--i}) shows the difference of means for the total time variable ($T_\mathrm{total}$). There is no overlap between the HDI and the ROPE for all pairwise comparisons between methods, except for an overlap of 9\% between \textbf{focus} and \textbf{no~path}. The median value, 45.4 seconds, of the difference of means ($\mu_\mathrm{focus}-\mu_\mathrm{NoPath}$) shows that our method is 8.17\% more time-consuming than the \textbf{no~path} baseline. In contrast, the \textbf{sampling} method considerably increases the computation time compared to the \textbf{no~path} baseline, as the median of the difference of means is 278 seconds, which corresponds to 50\% additional time. Finally, when comparing the \textbf{focus} and \textbf{sampling} strategies, the median of difference of means is around -232 seconds, meaning that the our strategy is 27.8\% times faster than the state of the art. 
\begingroup
\setlength{\tabcolsep}{4pt} 
\begin{table*}[t]

\caption{Mean (sd) object reconstruction performance in terms of coverage, AUC, entropy, and total time ($T_\mathrm{total}=T_\mathrm{comp}+T_\mathrm{exec}$) along with computation time ($T_\mathrm{comp}$) and execution time ($T_\mathrm{exec}$) for varying camera FoV and model resolution.}
\label{tab:cov_ent_time_sensitivity}
\begin{center}
\begin{tabular}{m{2cm}m{0.95cm}m{0.95cm}m{0.95cm}|m{0.1cm}m{0.95cm}m{0.95cm}m{0.95cm}|m{0.1cm}m{0.95cm}m{0.95cm}m{0.95cm}}
&\multicolumn{3}{c}{$\mathrm{FoV}=(74^\circ,60^\circ)$}&&\multicolumn{3}{c}{$\mathrm{FoV}=(60^\circ,48^\circ)$}&&\multicolumn{3}{c}{$\mathrm{FoV}=(45^\circ,36^\circ)$}\\
&\multicolumn{3}{c}{$r=0.12~m$}&&\multicolumn{3}{c}{$r=0.03~m$}&&\multicolumn{3}{c}{$r=0.03~m$}\\
\hline
\hline
 &Focus&Sampling&No~path&&Focus&Sampling&No~path&&Focus&Sampling&No~path\\
\hline
Coverage&\makecell{0.826\\(0.093)}&\makecell{\underline{\textbf{0.828}}\\(\underline{\textbf{0.092}})}&\makecell{0.794\\(0.103)}&&\makecell{0.806\\(\underline{\textbf{0.1}})}&\makecell{\underline{\textbf{0.814}}\\(\underline{\textbf{0.1}})}&\makecell{0.76\\(0.113)}&&\makecell{\underline{\textbf{0.78}}\\(0.107)}&\makecell{\underline{\textbf{0.78}}\\(\underline{\textbf{0.106}})}&\makecell{0.701\\(0.127)}\\
AUC&\makecell{0.679\\(0.09)}&\makecell{\underline{\textbf{0.683}}\\(\underline{\textbf{0.085}})}&\makecell{0.61\\(0.103)}&&\makecell{0.657\\(0.09)}&\makecell{\underline{\textbf{0.665}}\\(\underline{\textbf{0.089}})}&\makecell{0.561\\(0.108)}&&\makecell{\underline{\textbf{0.62}}\\(\underline{\textbf{0.099}})}&\makecell{0.616\\(0.1)}&\makecell{0.486\\(0.117)}\\
Entropy ($\times 10^3$)&\makecell{\underline{\textbf{1825.21}}\\(121.096)}&\makecell{1848.14\\(\underline{\textbf{117.952}})}&\makecell{1989.85\\(155.32)}&&\makecell{\underline{\textbf{1880.64}}\\(\underline{\textbf{133.466}})}&\makecell{1911.60\\(146.925)}&\makecell{2109.81\\(168.758)}&&\makecell{\underline{\textbf{2024.99}}\\(\underline{\textbf{133.559}})}&\makecell{2080.97\\(143.440)}&\makecell{2252.76\\(161.614)}\\
\hline
$T_\mathrm{comp}$ (s)&\makecell{\underline{\textbf{3.27}}\\(\underline{\textbf{1.02}})}&\makecell{63.31\\(28.77)}&\quad\textemdash&&\makecell{\underline{\textbf{11.81}}\\(\underline{\textbf{2.70}})}&\makecell{136.59\\(53.16)}&\quad\textemdash&&\makecell{\underline{\textbf{12.57}}\\(\underline{\textbf{2.65}})}&\makecell{77.98\\(26.08)}&\quad\textemdash\\
$T_\mathrm{exec}$ (s)&\makecell{577.69\\(164.07)}&\makecell{639.18\\(215.67)}&\makecell{\underline{\textbf{555.37}}\\(\underline{\textbf{81.65}})}&&\makecell{588.18\\(105.43)}&\makecell{626.52\\(139.45)}&\makecell{\underline{\textbf{547.66}}\\(\underline{\textbf{86.0}})}&&\makecell{604.31\\(101.69)}&\makecell{637.39\\(118.17)}&\makecell{\underline{\textbf{547.76}}\\(\underline{\textbf{87.25}})}\\
$T_\mathrm{total}$ (s)&\makecell{580.96\\(164.89)}&\makecell{702.49\\(241.78)}&\makecell{\underline{\textbf{555.37}}\\(\underline{\textbf{81.65}})}& &\makecell{599.99\\(107.09)}&\makecell{763.11\\(188.64)}&\makecell{\underline{\textbf{547.66}}\\(\underline{\textbf{86.0}})}&&\makecell{616.88\\(103.31)}&\makecell{715.37\\(138.96)}&\makecell{\underline{\textbf{547.76}}\\(\underline{\textbf{87.25}})}\\
\hline
\end{tabular}
\end{center}

\end{table*}
\endgroup

\subsection{Sensitivity Analysis}
A sensitivity analysis was conducted to evaluate the influence of camera FoV and map resolution on the algorithms' performance. To this end, we performed simulations containing 114 objects for two additional FoV angles,  ($60^{\circ}$, $48^{\circ}$) and ($45^{\circ}$, $36^{\circ}$), and a lower resolution of $\unit[0.12]{m}$. The model resolution to calculate the comparison metrics uses the original resolution of $\unit[0.03]{m}$, with the lower resolution of $\unit[0.12]{m}$  only being used when evaluating views for the sampling-based method and calculating the focus point for our method.

Table~\ref{tab:cov_ent_time_sensitivity} presents the coverage, entropy, and total time ($T_\mathrm{total}$) along with computation time ($T_\mathrm{comp}$) and execution time ($T_\mathrm{exec}$) for varying settings. Considering coverage under different settings, using a lower resolution ($\unit[0.12]{m}$ instead of $\unit[0.03]{m}$) while maintaining the same camera FoV of ($74^\circ,60^\circ$) results in a slightly reduced coverage for both \textbf{focus} (from 0.833 to 0.826, or 0.7\%) and \textbf{sampling} (from 0.84 to 0.828, or 1.2\%) methods. In terms of entropy, both \textbf{focus} and \textbf{sampling} demonstrate higher final entropy compared to the high resolution setting. Specifically, the entropy for the \textbf{focus} method increases 1.61\%, from 1796.25 to 1825.21, and for the \textbf{sampling} method it increases 2.34\%, from 1805.75 to 1848.14. Finally, the total times for both methods decrease mainly due to the computation times, as shown in Table~\ref{tab:cov_ent_time_sensitivity}. This is because the number of voxels evaluated along the rays is smaller when the map resolution is lower. Under this setting, the total times ($T_\mathrm{total}$) for our method and \textbf{sampling} are 580.96~s and 702.49~s, respectively. This means that our method is approximately 17.29\% faster in terms of the total time ($T_\mathrm{total}$) spent across all NBVs. By looking at the computation times ($T_\mathrm{comp}$) for this setting, one can see that our method is approximately 19.36 times faster than \textbf{sampling} method (i.e., $63.31/3.27$). 

In reduced FoV settings, the final coverage for both \textbf{focus} and \textbf{sampling} decreases, while the entropy increases, as shown in Table~\ref{tab:cov_ent_time_sensitivity}. This is because a reduced camera FoV measures a smaller volume at each observation, implying less information gathered about the object and its bounding box. Regarding average computation time ($T_\mathrm{comp}$), the time required by our method remains nearly constant for different camera FoVs ($\unit[11.81]{s}$ and $\unit[12.57]{s}$, from larger to smaller FoV), whereas it decreases as the camera FoV becomes smaller for the sampling method ($\unit[136.59]{s}$ and $\unit[77.98]{s}$, from larger to smaller FoV). This is because our method employs a fixed expanded FoV ($90^\circ, 90^\circ$) regardless of the actual camera FoV. In contrast, the sampling method evaluates candidate views using the real camera FoV. Under these settings, our method is 11.56 and 6.2 times faster than \textbf{sampling} for $(60^\circ,48^\circ)$ and $(45^\circ,36^\circ)$, respectively, in terms of computation time ($T_\mathrm{comp}$). Considering the total time ($T_\mathrm{total}$), our method is 21.37\% and 13.76\% faster than \textbf{sampling} for $(60^\circ,48^\circ)$ and $(45^\circ,36^\circ)$, respectively.

The total time ($T_\mathrm{total}$) for \textbf{no~path} method does not depend on varying FoV and model resolution since it does not use any strategy to collect information along the path. However, coverage, entropy, and AUC values decrease as the FoV becomes smaller, because a reduced camera FoV captures a smaller volume at each observation. Finally, although \textbf{no~path} is the fastest method in terms of total time ($T_\mathrm{total}$), its performance is worse than \textbf{focus} and \textbf{sampling} regarding coverage, AUC and entropy for any of the evaluated FoVs and model resolutions.

\subsection{Ablation Study}

An ablation study has been conducted to illustrate the individual effects of the focus point strategy and the visibility constraint. To examine the isolated effects, we considered $\mathrm{FoV}=(74^\circ,60^\circ)$, model resolution $r=\unit[0.03]{m}$, and three conditions: (1) the visibility constraint is disabled and the focus point calculation is enabled; (2) the visibility constraint is enabled and the focus point calculation is disabled; and (3) both the visibility constraint and focus point calculation are enabled. We then repeated the previous simulations for 114 objects across all these three conditions.

The first case, where the focus point is calculated using the proposed method but the visibility constraint is not included in the optimization problem~\ref{eq:control-law}, is analogous to the \textbf{no~path} strategy. The focus point is enforced through the visibility constraint, which is only effective when incorporated into the control law~\ref{eq:control-law} (that is, the visibility constraint is necessary to focus on a particular point). Therefore, in this case, although the focus point is calculated, the robot moves to the NBV without keeping the focus point within the camera FoV, analogous to the \textbf{no~path} strategy. 

In the second case, the visibility constraint is included in the control law~\ref{eq:control-law}, but the focus point method proposed in Section~\ref{sec:focus_point_calculation} is not used to estimate an informative focus point. Nonetheless, an arbitrary point is required to enforce the visibility constraint, as the constraint is based on the distance dynamics between a given reference point and the camera's FoV planes. Therefore, we used the center of the search space where the object is located as the reference point for the visibility constraint.

In the final case, we consider both the visibility constraint and the informative focus point calculation strategy proposed in Section~\ref{sec:focus_point_calculation}, where the robot keeps the calculated informative focus point within the camera FoV by enforcing the visibility constraint in the control law~\ref{eq:control-law}.

Table~\ref{tab:comparison_visibility} illustrates the object coverage, AUC, entropy and total time ($T_\mathrm{total}$) for three considered cases. Considering object coverage, when both the visibility constraint and informative focus point estimation are active, the object coverage improves by 3.9\% and 3.5\% and the entropy reduces by 9.72\% and 12.45\%  compared to using only focus point estimation and only the visibility constraint, respectively. Finally, using both strategies increases the total time ($T_\mathrm{total}$) by 8.12\% compared to using only the focus point strategy and by 8.88\% compared to using only the visibility constraint. Therefore, it is clear that using the visibility constraint with an informative focus point improves coverage, AUC, and entropy at the expense of a small increase in computation and execution time.

\begingroup
\setlength{\tabcolsep}{4pt} 
\begin{table}[h]
\caption{The mean ($\mu$) and standard deviation ($\sigma$) of object coverage, AUC, entropy, and total time ($T_\mathrm{total}$) are reported for three cases at the 10\textsuperscript{th} NBV: using only the focus point estimation strategy (\textbf{Only} Focus), using only the visibility constraint (\textbf{Only} Visibility), and using both strategies (\textbf{Both}).}
\label{tab:comparison_visibility}
\begin{center}

\begin{tabular}{m{1.9cm}m{1.95cm}m{1.95cm}m{1.95cm}}
 &\textbf{Only} Focus&\textbf{Only} Visibility&\textbf{Both}\\
\hline
\hline
&$\mu\ (\sigma)$&$\mu\ (\sigma)$&$\mu\ (\sigma)$\\
Coverage&0.794 (0.103)&0.798 (0.101)& \underline{\textbf{0.833}} (\underline{\textbf{0.093}})\\
AUC&0.613 (0.103)&0.622 (0.097)& \underline{\textbf{0.684}} (\underline{\textbf{0.083}})\\
Entropy ($\times 10^3$)&1989.85 (155.32)&2051.85 (148.2)& \underline{\textbf{1796.25}} (\underline{\textbf{128.95}})\\
 $T_\mathrm{total}$ (s)&555.37 (\underline{\textbf{81.65}})&\underline{\textbf{551.46}} (93.75)& 600.46 (126.72)\\
\hline
\end{tabular}

\end{center}
\end{table}

\endgroup

\begin{figure}[t]
      \centering
      \includegraphics[width=1\linewidth]{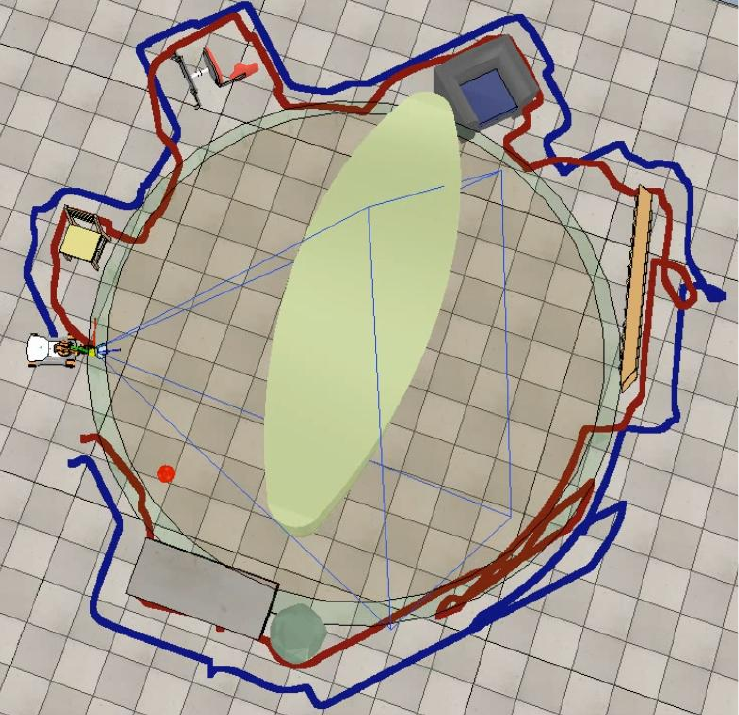}
      \caption{Illustration of collision avoidance for the robot base and end-effector. The base and end-effector paths are shown in blue and red, respectively.}
      \label{fig:robot_path_obstacles}
\end{figure}

\subsection{Robot Motion in Cluttered Environments}
We performed simulations for 20 objects in a cluttered environment with box- and cylinder-type obstacles, similar to Fig.~\ref{fig:robot_path_obstacles}, to demonstrate the method’s obstacle avoidance performance. In all runs, our algorithm reconstructs the object using informative focus points and visibility constraints while successfully avoiding obstacles for both the base and the end-effector. Fig.~\ref{fig:robot_path_obstacles} illustrates a typical path of the robot base and end-effector in a cluttered environment. Table~\ref{tab:comparison_with_without_obstacle} illustrates the results of our method in environments with and without obstacles. Remarkably, the reconstruction performance in cluttered environment is better than obstacle-free setting. Specifically, the object coverage increases by 1.8\%, from 0.844 to 0.862, and entropy decreases by 2.87\%, from 1770.51 to 1719.60. This is because the robot spends, on average, 56.85\% more time traveling in the cluttered environment, and the proposed method enables the robot to focus on informative regions as much as possible, even while avoiding the obstacles.
\begin{figure}[t]
      \centering
      \includegraphics[width=1\linewidth]{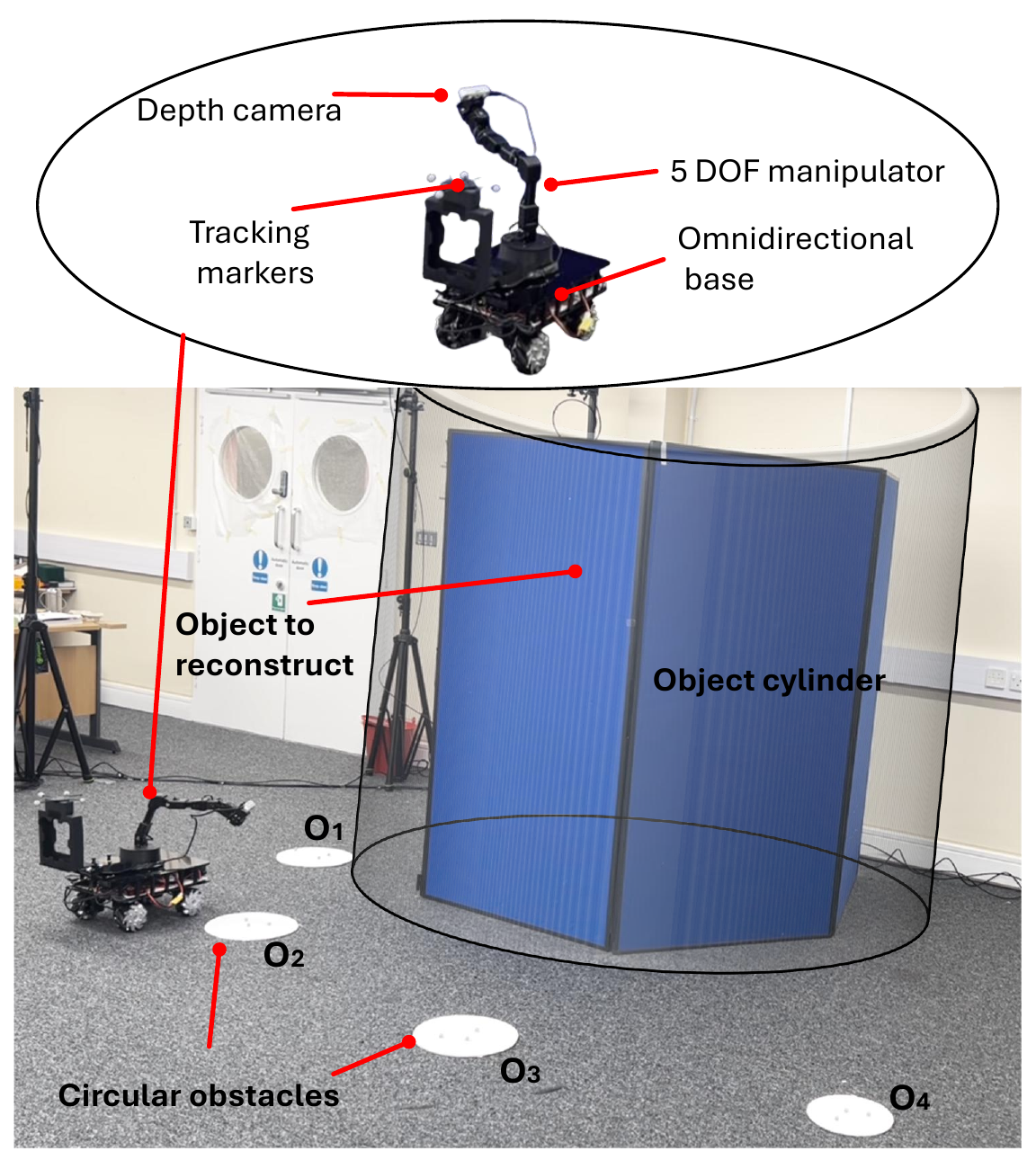}
      \caption{The first experimental setup consists of the object to be reconstructed, enclosed within a cylinder (2 m in diameter), along with four circular obstacles and an 8-DoF holonomic mobile manipulator. The robot base is enclosed by a cylinder with a diameter of 0.5~m, whereas obstacles $O_1,\ldots, O_3$ have a diameter of 0.3~m and $O_4$ has a diameter of 0.24~m.}
      \label{fig:real_world_setup}
\end{figure}
\begin{figure*}[htb!]
  \centering
  \begin{subfigure}[b]{0.65\columnwidth}
    \includegraphics[width=\linewidth]{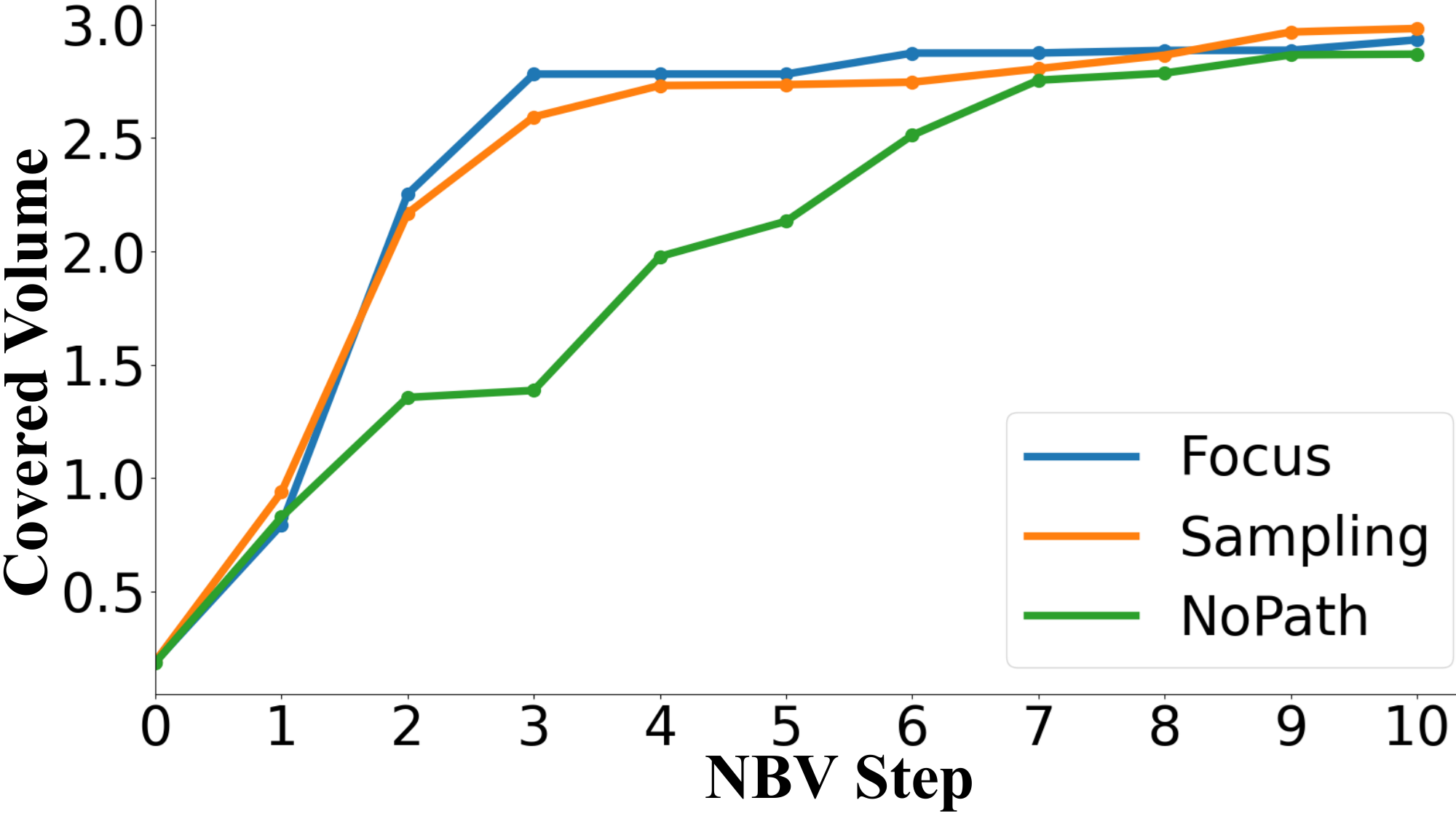}
    \caption{}
     
  \end{subfigure}
  \begin{subfigure}[b]{0.65\columnwidth}
    \includegraphics[width=\linewidth]{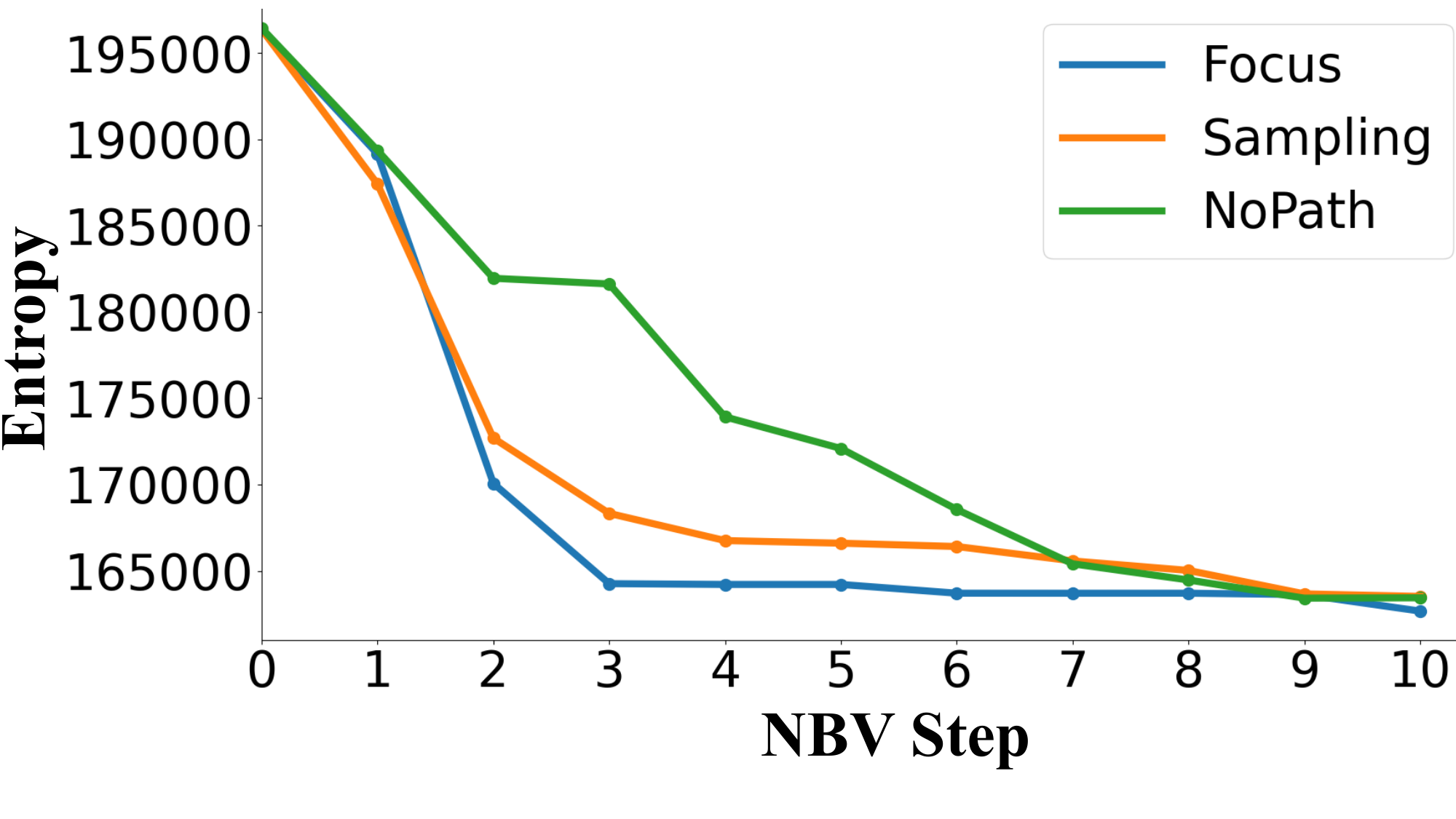}
    \caption{}
     
  \end{subfigure}
  \begin{subfigure}[b]{0.65\columnwidth}
    \includegraphics[width=\linewidth]{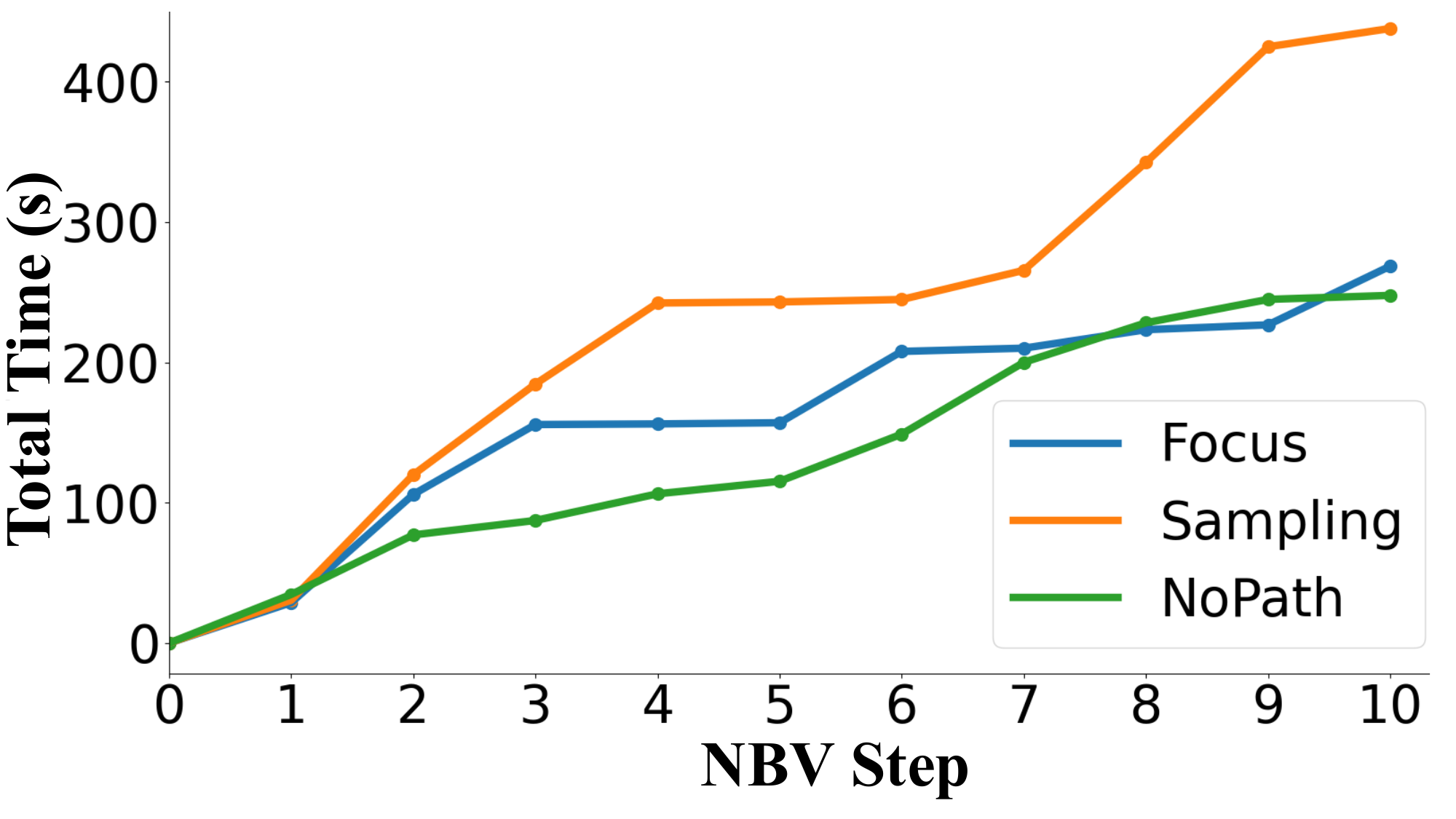}
    \caption{}
   
  \end{subfigure}
  \caption{Covered volume, entropy, and total time for consecutive NBV steps in the real-world experiment with the holonomic wheeled mobile manipulator for a single run. Our method (\textbf{focus}) and the \textbf{sampling} method both demonstrate similar covered volume and entropy performance, outperforming the \textbf{no path} method except for final NBVs where all three methods converge to similar values. Additionally, our method requires significantly less time than the \textbf{sampling} method.}
  \label{fig:real_robot_data}
\end{figure*}

\begingroup
\setlength{\tabcolsep}{4pt} 
\begin{table}[h]
\caption{The mean ($\mu$) and standard deviation ($\sigma$) of object coverage, AUC, entropy, and total time ($T_\mathrm{total}$) are reported at the 10\textsuperscript{th} NBV for 20 objects using our method in environments \textbf{with} and \textbf{without} obstacles. }
\label{tab:comparison_with_without_obstacle}
\begin{center}

\begin{tabular}{m{1.9cm}m{2.3cm}m{1.95cm}}
 &\textbf{Without} Obstacles&\textbf{With} Obstacles\\
\hline
\hline
&$\mu\ (\sigma)$&$\mu\ (\sigma)$\\
Coverage&0.844 (0.078)&\underline{\textbf{0.862}} (\underline{\textbf{0.069}})\\
AUC&0.699 (0.075)&\underline{\textbf{0.71}} (\underline{\textbf{0.068}})\\
Entropy ($\times 10^3$)&1770.51 (71.141)&\underline{\textbf{1719.60}} (\underline{\textbf{67.974}})\\
 $T_\mathrm{total}$ (s)&\underline{\textbf{600.93}} (\underline{\textbf{139.01}})&942.57 (193.24)\\
\hline
\end{tabular}

\end{center}
\end{table}

\endgroup

\subsection{Experimental Results}
We performed two experiments: (1) using a holonomic wheeled mobile manipulator to compare the \textbf{focus} (ours), \textbf{sampling}, and \textbf{no~path} methods in a real-world setup; and (2) using a legged manipulator to demonstrate the generality of the proposed focus point estimation and visibility constraint method.
\subsubsection{Holonomic Wheeled Mobile Manipulator}\label{experiment:mobile-manipulator}
In the first real-world experiment, we employed the setup shown in Fig.~\ref{fig:real_world_setup}, which included a custom omnidirectional mobile base equipped with a 5-DOF arm\footnote{\href{https://docs.trossenrobotics.com/interbotix_xsarms_docs/specifications/rx150.html}{Trossen Robotics Reactor X-150}} and an Intel RealSense D435 depth camera mounted on the end-effector. The camera provided depth data for volumetric mapping using the OctoMap library \cite{Ref:Octomap}. An NVIDIA Orin Nano\footnote{\href{https://developer.nvidia.com/embedded/learn/get-started-jetson-orin-nano-devkit}{NVIDIA Orin Nano}} onboard computer handled the robot motion control using DQ Robotics \cite{Ref:DQ} library and real-time volumetric model updates using the OctoMap library \cite{Ref:Octomap}, while a remote computer (the same one used in simulations) selected NBVs and focus points.

The robot localization was done using an OptiTrack \footnote{\href{https://www.optitrack.com}{https://www.optitrack.com}} motion capture system, with communication between the robot, remote computer, and OptiTrack established through ROS Noetic over Wi-fi network.

Due to workspace constraints and occlusions caused by the large reconstruction object, only 100 of the 200 predefined search-space views were feasible. The remaining ones were inaccessible because the object was positioned near a wall to accommodate the motion capture system and the robot’s marker. To increase task complexity, we placed four circular obstacles arbitrarily along the robot’s path.

We measured the same metrics used in the simulation, except for object coverage. Since we did not have access to a ground-truth model for the reconstructed object in the real-world setup, we instead used the covered (known) volume\textemdash i.e., the sum of occupied and free volumes\textemdash in place of object coverage. This volume is computed from the voxel size: with a voxel resolution $r = 0.03~\mathrm{m}$, each voxel has a volume $v = (0.03)^3~\mathrm{m}^3$.

Considering the covered volume, Fig.~\ref{fig:real_robot_data}(a) shows that both the \textbf{sampling} method and our method (\textbf{focus}) increase the covered volume much faster than the \textbf{no~path} baseline. After the 7\textsuperscript{th} NBV, the covered volumes of all three methods reach similar values of $2.934~\mathrm{m}^3$, $2.983~\mathrm{m}^3$ and $2.870~\mathrm{m}^3$ for \textbf{focus}, \textbf{sampling} and \textbf{no path}, respectively.  A similar trend appears in the entropy metric (Fig.~\ref{fig:real_robot_data}(b)), as entropy reduction under the \textbf{focus} and \textbf{sampling} strategies is significantly faster in the early stages, but by the 10\textsuperscript{th} NBV, all methods reach entropy values that differ by less than 1\% ($162670.531 $, $163519.734$ and $163419.906$ for \textbf{focus}, \textbf{sampling} and \textbf{no path}, respectively).

These results are expected, since we consider only half of the search space, and 10 NBV steps suffice in this setting. Indeed, all metrics when using the \textbf{focus} and \textbf{sampling} strategies change little after the 3\textsuperscript{rd} NBV, while the \textbf{no~path} baseline continues to improve more significantly.

Finally, Fig.~\ref{fig:real_robot_data}(\emph{c}) indicates that our method (\textbf{focus}) requires significantly less computation time than the \textbf{sampling} method, without increasing the time compared to \textbf{no~path}. 

During the object model's 3D reconstruction, some phantom voxels were observed due to many factors. First, the motion tracking system's marker is attached to the robot base while the camera is attached to the end-effector of the arm. Therefore, the camera pose is calculated based on the base pose, measured with the motion tracking system, and the arm kinematics. Since we use a low-budget robotic arm, errors in the camera pose estimation with respect to the base arise, directly affecting the generated point cloud and octomap. Second, the depth camera has measurement errors, which are directly proportional to the measurement distance, also contributing to the emergence of phantom voxels.

It is important to emphasize that this setup is more realistic than attaching markers directly on the end-effector. In the absence of a motion capture system, the robot localization would be primarily calculated for the base, with the end-effector pose being calculated through forward kinematics, similar to our current setup.

\begin{figure*}[htb!]
     \centering
     \includegraphics[width=1\linewidth]{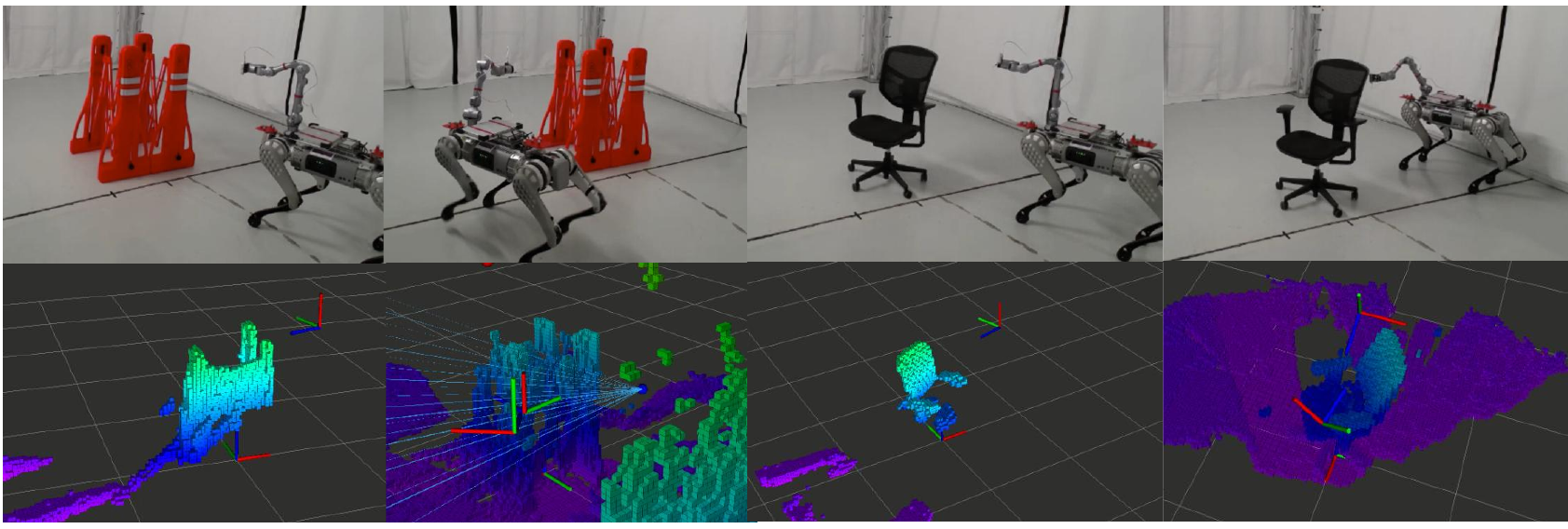}
     \caption{Snapshots of the reconstruction process for a barrier and a chair using a legged manipulator with a camera mounted on its end-effector.}
     \label{fig:unitree_b1z1_reconstruction}
\end{figure*}

\begin{figure}[h]
     \centering
     \includegraphics[width=1\linewidth]{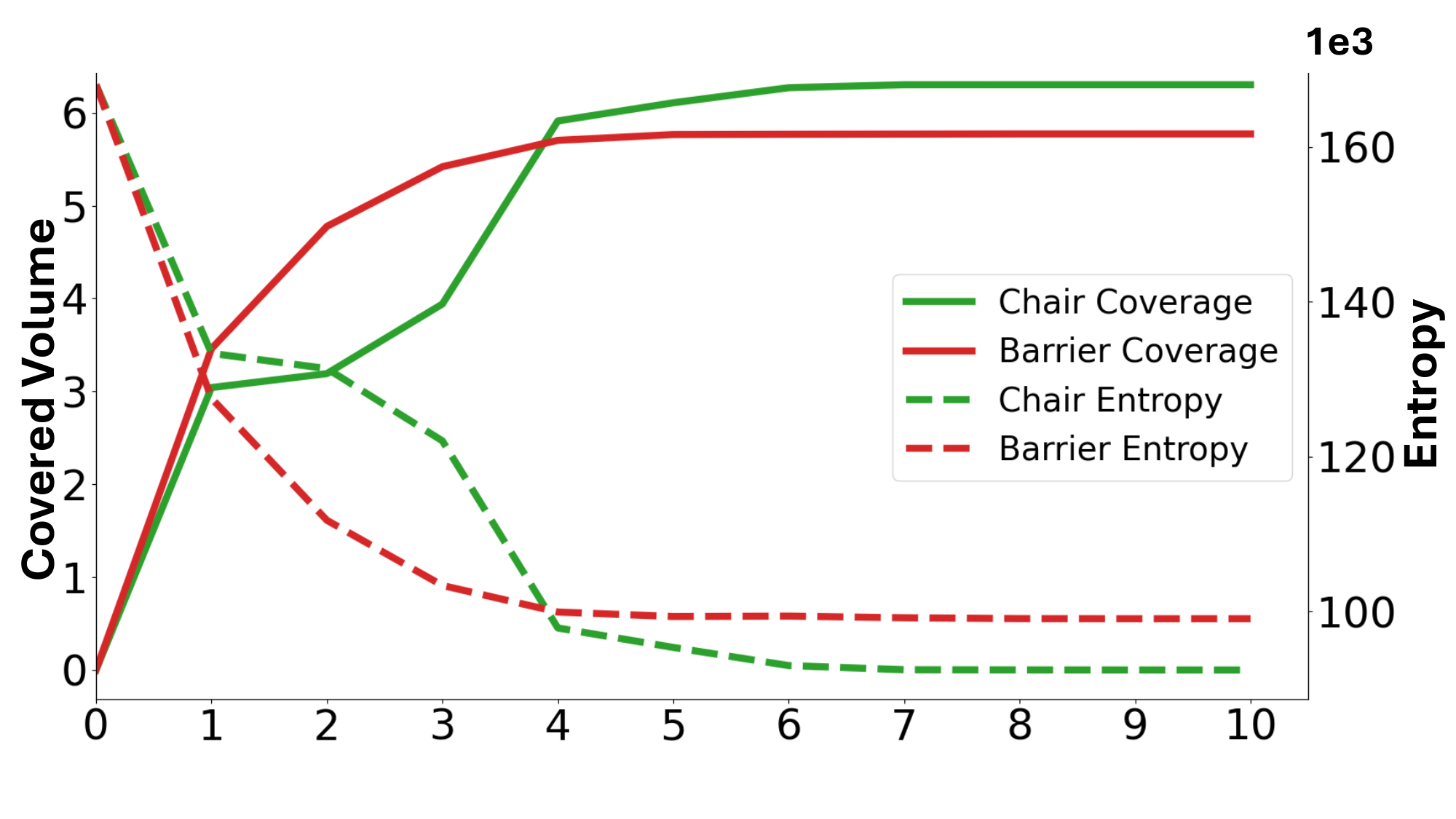}
     \caption{Covered volume and entropy for consecutive NBV steps in the real-world experiment with the legged manipulator when reconstructing a chair and a barrier.}
     \label{fig:unitree_real_robot_data}
\end{figure}

\subsubsection{Legged Manipulator}
To illustrate the generality of the proposed reconstruction-aware control strategy, we also conducted a second experiment using a Unitree B1 legged robot with a Unitree Z1 robotic arm. For this experiment, two objects were used: a barrier and a chair. Similarly to the experiment described in Section~\ref{experiment:mobile-manipulator}, a camera was mounted on its end-effector and the whole-body control law \eqref{eq:control-law} was used to coordinate the legged robot and the manipulator arm. The robot base is modeled using three spheres (front, middle, and rear), and the end-effector is modeled as a single sphere. To ensure safe operation, the following geometric constraints \cite{Ref:VFI_Marinho} were added: three point-to-line constraints to prevent collisions between the base and the virtual cylinder surrounding the object; one point-to-line constraint to avoid collisions between the end-effector and the same virtual cylinder; 12 point-to-plane constraints between the three base spheres and the four workspace walls; and four point-to-plane constraints between the end-effector sphere and the workspace walls.

Fig.~\ref{fig:unitree_b1z1_reconstruction} illustrates snapshots of the reconstruction process of an orange barrier and a chair. Fig.~\ref{fig:unitree_real_robot_data} demonstrates the corresponding covered volume, entropy, and total time for the barrier and chair objects.

Considering covered volume and entropy, the object reconstruction reaches a stable state around the 4\textsuperscript{th} NBV for the barrier object and around the 6\textsuperscript{th} NBV for the chair. At the final NBV (10\textsuperscript{th}), the chair reconstruction reaches lower entropy and higher covered volume. This outcome is expected since the chair object has a smaller interior volume that cannot be measured due to the occupied voxels around it, whereas the barrier object has a gap in the middle and the occupied voxels prevent the robot from taking measurements from those residual regions. The total reconstruction time is $\unit[473]{s}$ for the chair and  $\unit[514]{s}$ for the barrier.

Considering the created models, a similar phantom voxel effect observed in the previous experiment was also experienced in this setup, because the robot gait causes oscillations in the end-effector that also affects measurements.

\section{Conclusion}\label{sec:conclusion}

As object reconstruction is an automated process performed by a robot, measurements between different NBVs have a significant impact on the final performance of the object reconstruction process. The existing methods for object reconstruction generally involve micro-aerial vehicles, which are not restricted by obstacles on the ground but have a limited payload and typically have a short operational time due to battery capacity limitations. Furthermore, these methods address the issue of getting measurements between different NBVs by using sampling-based planners, which are computationally expensive, especially for high-resolution model representations. 

To address some of those limitations, this paper has proposed a computationally efficient method to accelerate the object reconstruction step without sacrificing object coverage, increasing entropy, or relying on low-resolution model representations. We have employed a mobile manipulator, which increases the payload and operational time, for autonomous reconstruction and integrated coverage awareness into the robot control. Our method allows the robot to focus on informative regions by directing its view towards informative regions within the object's bounding box while moving to the NBVs. We have compared our method with a sampling-based strategy, similar to the state of the art, and the results have shown that there is no significant difference in the object coverage and entropy, while our method is 6.2 to 19.36 times faster in terms of computation time and reduces the total time the robot spends between views by 13.76\% to 27.9\%, depending on the camera FoV and model resolution.Furthermore, we conducted real‑world experiments with an 8‑DOF mobile manipulator and a legged manipulator to demonstrate the proposed method’s performance in practice. All code and experimental data are open source, which can be found on \textbf{\url{https://github.com/fthdrsn/NBV_object_recontruction}}.

The current system requires knowledge of the locations of both the robot and obstacles for obstacle avoidance, which is practical when the workspace is static and both a map and a global localization system are available. However, this can be a limitation in more unstructured environments, where the map must be frequently updated to account for changes in the workspace, and localization must be done using local information. To overcome these limitations, future work will focus on implementing the proposed framework with less restrictive assumptions, where obstacle positions are unknown to the robot, and the robot will rely on sensor measurements (e.g., lidar and onboard cameras) to recognize obstacles dynamically.

\begin{figure}[htb!]
      \centering
      \includegraphics[width=1\linewidth]{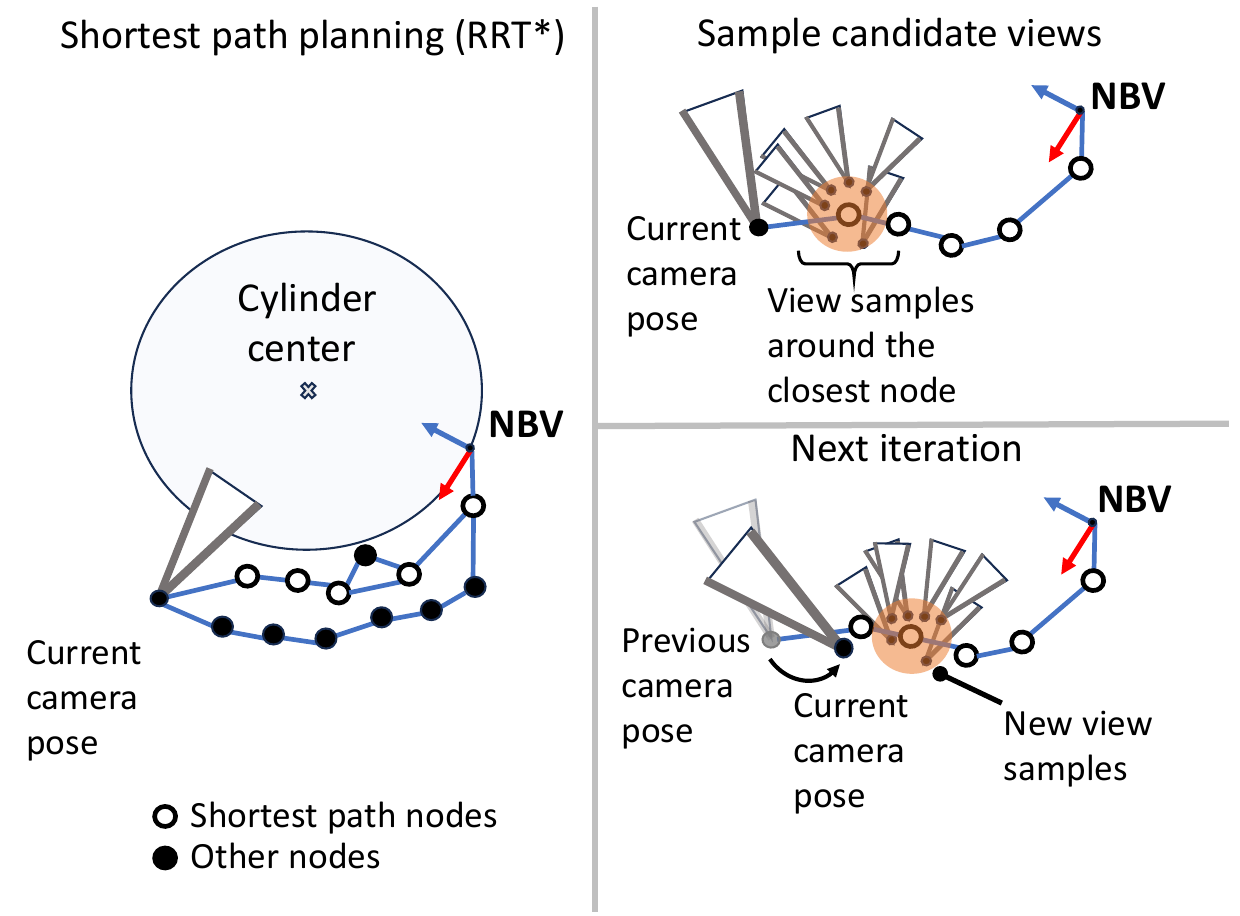}
      \caption{Planning an informative path with an RRT* sampling-based strategy.}
      \label{fig:sampling_planner}
\end{figure}

\appendix[Sampling-Based Informative Path Planning]
\label{sec:appendix}
We use a sampling-based approach to plan an informative path between NBVs as a baseline for comparing with our method described in Section~\ref{sec:focus_point_calculation}. Similar to \cite{Ref:Song_1, Ref:Song_2}, we find the shortest path between the current robot end-effector position and the NBV on a tree generated using $\text{RRT}^*$ \cite{Ref:RRT_Star} (Fig.~\ref{fig:sampling_planner}a). We then generate candidate views inside a sphere centered at the nodes of the shortest path (Fig.~\ref{fig:sampling_planner}b). This way, the generated views do not diverge significantly from the shortest path. The yaw and pitch angles are randomly sampled between $-\pi/6$ and $\pi/6$, and these angles are zero when the camera faces the center of the search space (i.e., the center of the object cylinder described in Section~~\ref{sec:focus_point_calculation}).  At each iteration, we only generate and evaluate samples around the closest node as the map updates after taking measurements. To evaluate each view, the total voxel entropy is calculated using \eqref{eq:voxel_entropy} inside the camera's FoV. After determining the best view around the closest node, the robot traverses to the selected view (Fig.~\ref{fig:sampling_planner}c), and the process is repeated until it reaches the NBV.

\bibliographystyle{IEEEtran}
\bibliography{IEEEabrv,Ref}

\end{document}